\documentclass{article}

\usepackage[preprint]{neurips_2024}

\usepackage{enumitem,xcolor}

\usepackage[utf8]{inputenc} 
\usepackage[T1]{fontenc}    
\usepackage{hyperref}       
\usepackage{url}            
\usepackage{booktabs}       
\usepackage{amsfonts}       
\usepackage{nicefrac}       
\usepackage{microtype}      
\usepackage{xcolor}         

\usepackage{url}            
\usepackage{booktabs}       
\usepackage{amsfonts}       
\usepackage{nicefrac}       
\usepackage{microtype}      
\usepackage{xcolor}         
\usepackage{tablefootnote}
\usepackage{amsmath,amsfonts,bm}
\usepackage{graphicx}
\usepackage{url}            
\usepackage{booktabs}       
\usepackage{nicefrac}       
\usepackage{microtype}      
\usepackage{wrapfig}
 \usepackage{array,multirow,graphicx}

\usepackage{breqn}
\usepackage{amsmath}

\usepackage{enumitem}
\usepackage[ruled,noend]{algorithm2e}
\usepackage{multirow}
\usepackage{tablefootnote}
\usepackage{multicol}
\usepackage{color}
\usepackage{colortbl}
\usepackage{xcolor}
\usepackage[capitalise]{cleveref}
\usepackage{mathtools}
\usepackage{wrapfig}
\usepackage{booktabs,subcaption,amsfonts,dcolumn}

\title{Multi-scale Image Super Resolution with a Single Auto-Regressive Model \vspace{-0.2cm}}

\author{%
  Enrique Sanchez\thanks{Equal contribution. Correspondence: \texttt{e.lozano@samsung.com}} \\ Samsung AI Center \\ Cambridge, UK \And Isma Hadji$^*$ \\ Samsung AI Center \\ Cambridge, UK 
  \And Adrian Bulat \\ Samsung AI Center \\ Cambridge, UK \And Christos Tzelepis\\ Samsung AI Center \\ Cambridge, UK \AND Brais Martinez \\ Samsung AI Center \\ Cambridge, UK \And Georgios Tzimiropoulos \\ Samsung AI Center \\ Cambridge, UK 
}

\begin{document}

\maketitle

\vspace{-0.5cm}
\begin{abstract}
    In this paper we tackle Image Super Resolution (ISR), using recent advances in Visual Auto-Regressive (VAR) modeling. VAR iteratively estimates the residual in latent space between gradually increasing image scales, a process referred to as next-scale prediction. Thus, the strong priors learned during pre-training align well with the downstream task (ISR). To our knowledge, only VARSR \cite{hu2025improvingautoregressivevisualgeneration} has exploited this synergy so far, showing promising results. However, due to the limitations of existing residual quantizers, VARSR works only at a fixed resolution, i.e. it fails to map intermediate outputs to the corresponding image scales. Additionally, it relies on a 1B transformer architecture (VAR-d24), and leverages a large-scale private dataset to achieve state-of-the-art results. We address these limitations through two novel components: a) a \textbf{Hierarchical Image Tokenization} approach with a multi-scale image tokenizer that progressively represents images at different scales while simultaneously enforcing token overlap across scales, and b) a \textbf{Direct Preference Optimization (DPO) regularization term} that, relying solely on the LR and HR tokenizations, encourages the transformer to produce the latter over the former. To the best of our knowledge, this is the first time a quantizer is trained to force semantically consistent residuals at different scales, and the first time that preference-based optimization is used to train a VAR. Using these two components, our model can denoise the LR image and super-resolve at half and full target upscale factors in a single forward pass.     Additionally, we achieve \textit{state-of-the-art results on ISR}, while using a small model (300M params vs ~1B params of VARSR), and without using external training data. Project site \url{https://github.com/saic-fi/ms_sr_var}. 
\end{abstract}

\section{Introduction}
\label{sec:intro}

Image Super-Resolution (ISR) is the task of generating a high-resolution image (HR) from a low-resolution (LR) version. Strong ISR methods typically rely on generative priors, either GAN-based~\cite{gansr,realesrgan_iccvw21} or diffusion-based~\cite{stablesr,yonos_eccv24,edge-sd-sr}, to enforce that the output image falls within the manifold of natural images. Recently, Visual Auto-Regressive (VAR) models have shown competitive performance for image generation, providing an alternative source of generative priors. We focus on the next-scale prediction variant due to its state-of-the-art image generation performance and its exceptional alignment between pre-training and downstream tasks. 

The only prior work is VARSR~\cite{qu2025visualautoregressivemodelingimage}, which adapts the general VAR framework for text-to-image (T2I~\cite{hart_iclr25,ma2024star,sun2024autoregressive}) or class-to-image (C2I~\cite{var, Infinity}) generation to the domain of ISR by conditioning the sequence generation on the features computed from the low-res image. This avoids the global sampling of diffusion models and the tendency to produce hallucinations of GAN-based methods, and provides state-of-the-art performance. However, VARSR also inherits the limitations of VAR, like requiring scale-specific models. In particular, most Auto-Regressive image generation methods rely on a Residual Quantizer (RQVAE~\cite{lee2022autoregressive}), a version of Vector Quantization (VQVAE~\cite{vector_quantization,van2017neural}) that offers a better fidelity-vocabulary size tradeoff than that of VQVAE, especially when applied to the next-scale prediction paradigm. However, applying Residual Quantization in a next-scale fashion does not guarantee that intermediate scales can be decoded into valid images (see \cref{fig:example} Top). This effect, while not being particularly important for the C2I or T2I cases, limits the derived ISR models to operate on fixed upsampling factors.

In this paper, we propose two technical contributions to advance VARSR. In particular, we first introduce a \textbf{Hierarchical Image Tokenization} approach in which the tokenization is applied sequentially to downsampled versions of the input image, while encouraging higher scales to reuse the tokens from lower scales. We show that by simply finetuning the vocabulary and decoder of an RQ-VAE trained for $512$ resolution using our hierarchical representation, the image tokenization produces a same-size sequence but with residuals that can be mapped to intermediate image resolutions (see \cref{fig:example} Bottom). Second, we propose to train the VAR using \textbf{a regularization term} that is \textbf{based on Direct Preference Optimization}~\cite{rafailov2023direct}, which drives the model to ``prefer'' the sequence of HR tokens over those of the LR. Besides super-resolving flexibly at different scale factors, our simple 310M model also surpasses VARSR 1B in the traditional ISR metrics in all benchmarks. Contrary to VARSR, \textit{we do \textbf{not} rely on external data to train our models}. Our main \textbf{contributions} can be summarized as follows:
\begin{itemize}
    \item We propose the first \textit{multi-scale} VAR-based approach for ISR by introducing a \textbf{Hierarchical RQ-VAE} that can decode intermediate scales to the image space, keeping the semantic information across scales consistent.
    \item We introduce a simple \textbf{DPO-based regularization term} that, contrary to VARSR, does not require collecting negative samples, thereby simplifying the training. 
\end{itemize}

\section{Background}
\label{sec:preliminaries}

\noindent \textbf{Image tokenization} A key component of VAR methods is the representation of images as a series of \textit{tokens}. An image is mapped to a latent representation using an autoencoder endowed with a discrete quantizer that maps continuous features to their closest representations from a fixed, learned vocabulary. The most common quantizer is Vector Quantization (\textit{VQ})~\cite{van2017neural,esser2021taming}). For an image $I \in \mathbb{R}^{3 \times H \times W}$, \textit{VQ} maps image features $\mathcal{E}(I) = {\bf Z} \in \mathbb{R}^{n_z \times h \times w}$ to an indexed set of $h \times w$ tokens from a learned codebook $\mathcal{V} = \{ {\bf r}_k\}_{k=1}^K \subset \mathbb{R}^{n_z}$. For a compression ratio $f\leq1$, we get $h \times w = fH \times fW$. The quantization $\mathcal{Q}$ of a feature vector ${\bf z}_k \in \mathbb{R}^{n_z}$ is described as 
\begin{equation}
\label{eq:quantization}
{\bf r}_k = \mathcal{Q}({\bf z}_k) = \operatorname{arg \, min}_{{\bf r}_{l} \in \mathcal{V}} \| {\bf r}_{l} - {\bf z}_k \| .   
\end{equation}
The model, consisting of an encoder $\mathcal{E}$, vocabulary $\mathcal{V}$, and decoder $\mathcal{D}$, is learned using a combination of image-reconstruction losses (e.g. $\ell_1$, perceptual, GAN) and a commitment loss~\cite{van2017neural}. The tokenization of an image is then the process of representing it as a sequence of discrete vectors $\{{\bf r}_1, ..., {\bf r}_{h \times w}\} = \mathcal{Q}(\mathcal{E}( \cdot ))$, which are then re-arranged as feature map $\mathbf{R} \in \mathbb{R}^{n_z \times h \times w}$ before being passed to the decoder to reconstruct the image $\tilde{I} = \mathcal{D}(\mathbf{R})$. Representing images as a \textit{sequence} of tokens is the key component behind the success of AR models. 

Residual Quantization (\textit{RQ}~\cite{lee2022autoregressive}) decomposes the latent ${\bf z}$ into a series of $L$ entries ${\bf r}_{l=1}^L$, such that $\sum_{l=1}^L {\bf r}_{l} \approx {\bf z}$, and $| {\bf z} - \sum_{l=1}^{L''} {\bf r}_{l} | \geq | {\bf z} - \sum_{l=1}^{L'} {\bf r}_{l} |$ for all $L'' < L' < L$. That is, \textit{RQ}  applies $\mathcal{Q}$ to the residuals $\delta_{L'} = {\bf z} - \sum_{l=1}^{L'} {\bf r}_{l}$, with ${\bf r}_1 = \mathcal{Q}({\bf z})$, and ${\bf r}_l = \mathcal{Q}(\delta_{l})$, such that the cumulative vectors approach the latent vector ${\bf z}$. Similarly to \textit{VQ}, \textit{RQ} is applied to each pixel of a latent representation ${\bf Z} \in \mathbb{R}^{n_z \times h \times w}$. While \textit{RQ} results in longer sequences for image tokenization, it comes with a smaller vocabulary size, facilitating the generation of better image sequences.

VAR~\cite{var} modifies the above formulation by making the residual $\Delta_{L'} = {\bf Z} - \sum_{l=1}^{L'} {\bf R}_l$ at each level $L'$ be quantized using an increasing number of tokens $n_{L'}$. That is, each level $L'$ is quantized using a number of tokens $n_{L'} \leq h \times w$. To do so, the quantization is done on each spatial location of $\Delta_{L'}$ after being downsampled to the target resolution $h' \times w'$. The map ${\bf R}_{L'} \in \mathbb{R}^{n_z \times h' \times w'}$ is upsampled back to the native resolution $h \times w$. Using this hierarchical decomposition, one can represent an image with shorter sequences; each level is now referred to as \textit{scales}. During image generation, rather than having a raster generation, the whole set of tokens corresponding to a given scale is produced using full attention. 

\textbf{Next-scale AR models} Recent approaches for AR-based image generation rely on the notion of next-scale prediction. An image is represented as a series of discrete, ordered tokens $\{z_{i,j}\}$ with $i$ representing the scale and $j$ representing the position within the scale. The goal of the AR model is to represent the conditional probability of a sequence as:
\begin{equation}
\label{eq:ar_formula}
p(z_{1,1}, \cdots, z_{L,n_L}) = \prod_{i=1}^L p( z_{i,1}, \cdots, z_{i,n_i} | z_{1,1} , \cdots , z_{i-1, n_{i-1}}, c).
\end{equation}
In \cref{eq:ar_formula}, $c$ represents the conditioning: in the text-to-image generation paradigm, $c$ is a text embedding, whereas in the class-to-image generation paradigm, $c$ represents a class number. Typically, an embedding representing $c$ is prepended to the sequence of tokens, so at inference the next-scale prediction starts from the target prompt or class.

\textbf{ISR using VAR} To the best of our knowledge, VARSR~\cite{qu2025visualautoregressivemodelingimage} is the first and only method thus far to introduce VAR to the ISR problem. VARSR applies the above formulation for super resolution using VAR as the main building block and making the conditioning $c$ represent the features computed on the input, low-res image. To ensure high-fidelity results, VARSR also introduces a new form of RoPE embeddings~\cite{su_roformer,ma2024star,hart_iclr25}, as well as an Image-based Classifier-free Guidance (CFG) training regularization, using a large pool of negative samples. While delivering new state-of-the-art results, \textit{VARSR relies on a large dataset of 4 million images gathered across 3k categories, as well as 50k negative samples from image quality assessment datasets}. Contrary to VARSR, we develop a DPO-based regularization term, thereby delivering new state-of-the-art results while training on standard ISR benchmarks. More importantly, we introduce hierarchical tokenization, enabling multi-scale ISR with the same model, thereby taking full advantage of the VAR framework.

\section{Method}
\begin{figure}[t]
    \centering
    \includegraphics[width=0.92\linewidth]{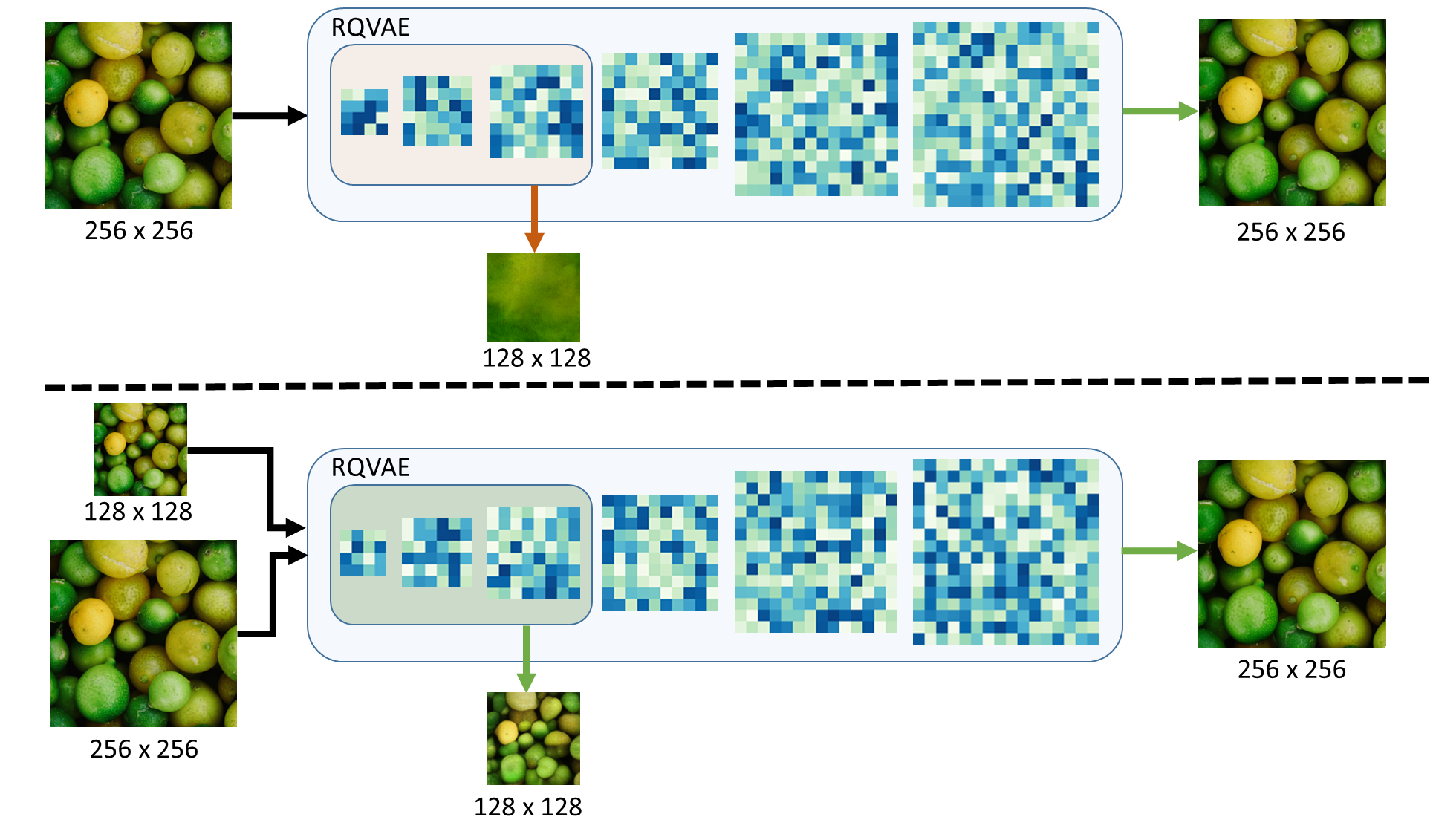}
    \caption{\textbf{Top}: we represent the quantization of an input image (up to $L=6$ scales for clarity of visualization). We observe that reconstructing the image using all 6 residuals leads to perfect reconstruction. However, mapping the residuals up to the first three scales does not result in the desired $2\times$-downsampled version of the input image, i.e., \textit{the initial scales do not convey scale-wise semantic information of the encoded image}. \textbf{Bottom}: Our proposed Hierarchical VQVAE Tokenization \cref{alg:enc} tackles the quantization in a hierarchically, enforcing the first set of tokens to be shared across both resolutions, thereby allowing reconstruction at both scales. }
    \label{fig:example}
    \vspace{-15pt}
\end{figure}

\paragraph{Motivation:} Despite the appealing properties of approaching ISR as an AR problem, we identify and aim to address the following two main shortcomings:
\begin{enumerate}
\item[\color{red} \textbf{a})] AR methods using a standard RQ-VAE fail to represent intermediate scales, as all residuals are needed to decode the final-scale image. Let $f \leq 1$ be the compression factor of the autoencoder. An input image of size $H \times W$ will produce a feature map $h \times w = fH \times fW$. In convolutional autoencoders, one can compress higher (or lower) resolution images using the same architecture whilst keeping $f$. In that sense, if an image is represented as a sequence of scales $\{z_{i,j}\}_{i=1, \cdots L, j=1 \cdots n_i}$, one would desire that the residuals up to $L' < L$ would lead the decoder to produce an image $H' \times W' = h_{L'}/f \times w_{L'}/f$. However, it is not guaranteed (and typically not true) that early residuals encode the semantic representation of the image, while additional scales add high-frequency details (see \cref{fig:example} (top) for a visual example). This lack of guarantees means that if a model has been trained to produce $4 \times$ upsample (i.e., $f=0.25$), it is not expected that the output at the scale corresponding to $2 \times$ ($f=0.5$) will indeed correspond to the $2 \times$-upsampled image. We address this shortcoming with our Hierarchical Image Tokenization.

\item[\color{red} \textbf{b})] In addition to not being able to produce reliable intermediate results, we observe that VARSR, contrary to most extant ISR work, needs additional annotated images (i.e., positive and negative samples) to use in Image-based Classifier-Free Guidance to help approach the desired HR targets. We address this shortcoming with our DPO-based regularization loss, which results in superior performance while only using the standard ISR training datasets.

\end{enumerate}

\paragraph{Hierarchical RQ-VAE} To address point {\color{red} \textbf{a})} we propose to partition the scale-based quantization by progressively encoding increasingly larger resolutions of the input image. We start by defining $s \leq 1$ as the target scale measured with respect to the feature dimensions $h \times w$ of the image to be quantized, $I \in \mathbb{R}^{3 \times H \times W}$. In other words, while $h,w = fHf, fWf$ is a fixed ratio, $s$ is the target scale in the feature space. We further define $\rho_L = (h, w)$ as the resolution of the feature map computed at the native resolution $H \times W$. We use the subscript $L$ to match the number of residual levels with the feature dimensions at the native resolution. Any further upscaling of the latent map would result in higher-resolution images. We define $\mathcal{E}(I)=\mathbf{Z} \in \mathbb{R}^{n_z \times h \times w}$ as the corresponding image features, and $\mathbf{Z}_{s} \in \mathbb{R}^{n_z \times sh \times sw}$ as the features corresponding to the downsampled image $I_{s} \in \mathbb{R}^{3 \times sH \times sW}$. Starting from the smallest scale, we propose to tokenize the residuals for each scale until the dimension of the residual surpasses the dimension of the corresponding features for that scale, i.e. until we reach step $i =  \mathrm{arg max}_k \, \rho_k \, | \, \rho_k < s_i \rho_L $. When moving into the next scale, the previous tokens are used to compute the residuals up to the previous scale, and further quantization is performed for that scale. In other words, we partition the resolutions for the residual quantization in non-overlapping subsets corresponding to the target scales, quantize the residuals as if the target scale was the last one, and append the tokens to the list computed from previous scales. The tokenization of scale $s_1$ sets the starting point for the tokenization of scale $s_2$, and so forth. \cref{alg:enc} summarizes the proposed tokenization procedure (also illustrated in \cref{fig:example} (bottom)).

At the end of the tokenization, an image is represented by a sequence of tokens as: 
\begin{equation}
    z = \bigg \{\underbracket{ \Big\{ \underbracket{  \{ \underbracket{z_{1}, z_2, \dots }_{s_1}\} , \dots , z_{l}, z_{l+1}, \dots }_{s_2} \Big\} ,  \dots ,z_L }_{s_N}\bigg\}
\end{equation}
where $s_i$ shows the set of tokens that can be used to decode scale $i$. This new image tokenization, based on downsampled versions of itself, will then be used to prepare the ground-truth sequences for the VAR training. However, because current quantizers do not have a vocabulary that is semantically consistent across scales, we finetune the vocabulary and the decoder of the RQ-VAE to accommodate the multi-scale quantization. To do this finetuning, we incorporate a scale-specific decoder, and compute the standard reconstruction losses for each scale. We then keep the decoder frozen and update the vocabulary using the gradient of the $\ell_2$ distance between the encoder features at each of the scales and the embeddings resulting from the tokenization. 
    
    \begin{algorithm}[t]
    \caption{\small{~Hierarchical RQVAE Tokenization}} \label{alg:enc}
    
    {\textbf{Inputs: } $\{I_n\}_{n=1}^N$} \; 
    \textbf{Hyperparameters: } steps $L$, resolutions $(\rho_l)_{l=1}^{L}$ , target scales $(0,s_1, \cdots, s_N)$ \;
    
    \For {$n=1,\cdots,N$}
    {
    $\mathbf{Z}_n = \mathcal{E}(\text{interpolate}(I, s_i \rho_L))$ \\
    \For {$i=1, \cdots , \max_{k} \rho_k \leq s_i \rho_L$}
    {
    \If {$i < \max_{k} \rho_k \leq s_{i-1} \rho_L$}{ 
       $\mathbf{Z}_n = \mathbf{Z}_n - \phi(\text{interpolate}(\mathbf{R}_{n,i}))$
    }
    \Else {
    
    $(z_{n,i}, \mathbf{R}_{n,i}) = \mathcal{Q}(\text{interpolate}(\mathbf{Z}_n, \rho_i ))$\\
    $\mathbf{Z}_n = \mathbf{Z}_n - \phi(\text{interpolate}(\mathbf{R}_{n,i}, \rho_n))$ \\
    $z \leftarrow [z; z_{n,i}]$ \\
    $r \leftarrow [r; \mathbf{R}_{n,i}]$
    }
    }
    
    }
    \textbf{Return: } multi-scale tokens $z$\;
    
  \end{algorithm}

\paragraph{DPO regularization of VAR training:} To address point {\color{red} \textbf{b})}, we observe that the model can be prone to bypass the difficulty of the ISR problem by directly predicting the LR tokens instead of the HR ones. This problem is also enhanced by the fact that there is an expected overlap between the tokens in both the LR and HR images. To overcome this, we enforce the model to not only maximize the log-likelihood of the ground-truth tokens, but also penalize the model when the output tokens are closer to describing a simple bilinear upsampling of the LR image. This type of objective is similar to that of Direct Preference Optimization~\cite{rafailov2023direct}, where the network is trained to prioritize a \textit{preferred} sequence over the \textit{non-preferred} one. Because we are not given a reference model, we simply choose a simpler regularization term of the form:
\begin{equation}
\mathcal{L}_{DPO} = - \log \sigma \left( \beta \log \frac{p(z_{HR})}{p(z_{LR})} \right)
\end{equation}
To compute $p(z_{LR})$, we upsample the low-res degraded image to the largest scale and tokenize it following \cref{alg:enc}. We define then $p(z_{LR}) = \prod_{z_i \in z_{LR}} \left(\exp{o_{z_i}}/(\sum_v \exp{o_v})\right)$ with $o = (o_1, \dots, o_K)$ the output logits from VAR. 
To the best of our knowledge, this is the first time a DPO-based regularization term is introduced in the context of AR image generation. 

\paragraph{Training and model details} We train the VAR with our Hierarchical Image Tokenization approach to represent the ground truth. We use a combination of a cross-entropy loss and our proposed DPO regularization term with equal weights for the two losses. 

To allow the VAR to handle multiple resolutions, we use an overparameterized learnable positional embedding which is downsampled to each of the $l$ resolutions $\rho_l$. Additionally, instead of using a ControlNet~\cite{zhang2023adding} to encode the LR image as conditioning to the VAR, as in VARSR, we directly use the RQVAE encoder features. We bilinearly upsample the LR to the target resolution before encoding it, and use the corresponding features as input to the VAR. 

Finally, we train our models using standard RealESRGAN degradations~\cite{realesrgan_iccvw21}. However, with a small probability $p\leq0.25$, we only apply a bilinear downsample of the HR image. We use a three-class embedding to represent degraded images, non-degraded images, and a class-free option for both degraded and non-degraded images. At inference time, setting the class to $1$ will result in small corrections to a simple bilinear upsample of the LR image. Conversely, using class $0$ results in the best performance for the upsampling of LR images under strong degradation.

\paragraph{Limitations:} By partitioning the number of steps into increasing scales, we are forcing the first levels to encode all information regarding scale $s_1$. This leaves a smaller number of steps to encode the second scale, and so forth. A full sequence of $L$ steps dedicated to representing a single resolution $H \times W$ is by definition stronger than if the same number of steps is required to encode the feature maps at lower resolutions as well. For example, if $L=10$ and we use the first five steps to encode scale, $s_1 = 0.5$, the reconstruction at $H' \times W' = H/2 \times W/2$ will be better than that of existing RQVAEs, but will incur a small image degradation at the full resolution $H \times W$. Therefore, there is a tradeoff between the number of scales encoded by the sequence $L$ and the quality of the decoded images. In this paper, we follow prior work and use $L=10$, and we partition the number of steps into three scales $s = (0.25, 0.5, 1)$. Increasing $L$ can however minimize the image degradation, albeit at the expense of a higher computational budget. 
\section{Experiments}
\subsection{Implementation details}
To follow prior work and for a fair comparison, we use the standard $4 \times$ ISR setting where the input LR images are of $128 \times 128$ resolution, while the target resolution is $512 \times 512$. Throughout the experiments, we use three scales $s = (0.25, 0.5, 1)$, corresponding to $1 \times$, $2 \times$ and $4 \times$ upsampling, respectively. The first scale can be considered as a simple image denoising.
\paragraph{Hierarchical RQVAE.} The compression factor of the encoder is $f=0.25$. To align with prior work on VAR, we use $L = 10$ steps, with resolutions $\rho_l = (4,6,8,10,14,16,20,24,28,32)$. Steps $l=1\dots3$ are mapped to the $128 \times 128$ output; steps $l=1\dots6$ result in the $256\times256$ outputs, and steps $l=1\dots10$ are used to perform the full $4\times$ upscale. Our RQVAE is initialized from the Switti checkpoint~\cite{switti}. To finetune the RQVAE we follow a hybrid approach akin to that of HART~\cite{hart_iclr25}, i.e., we randomly drop the quantization step and directly pass the encoder features to the decoder, with $50\%$ probability. We finetune our RQVAE vocabulary and decoders on the OpenImages~\cite{kuznetsova2020open} dataset for 25K iterations using AdamW~\cite{adamw} optimizer with batch size $384$, a learning rate of $0.00025$, and weight decay $0.05$. The loss is defined as $\mathcal{L}_{RQVAE}=\ell_2 + 5 \,\mathcal{L}_{perceptual}$, with $\mathcal{L}_{perceptual}$ being the standard LPIPs loss~\cite{zhang2018perceptual}. We train our RQVAE with 24 A100 GPUs, taking $\sim 24$ hours to complete. 

\paragraph{Hierarchical VAR.} The transformer follows the standard GPT-2 style~\cite{radford2019language}, adopted by VAR~\cite{var} and  VARSR~\cite{qu2025visualautoregressivemodelingimage}. However, contrary to VARSR, we use a transformer with only $16$ blocks, resulting in a 310M parameter transformer. 
Following VARSR, we initialize our model from the VAR d-16 official checkpoint. The model is trained using 24 A100 GPUs for 200 epochs, with a batch size of 384, learning rate of 1e-3, weight decay 0.005, and using an AdamW optimizer with betas $(0.9, 0.95)$. It takes $\sim 13$ hours for the training to be completed. To compute the LR features, we upsample the LR image to $512$ and encode it using the RQVAE encoder, producing a set of $1024$ conditioning tokens. The total number of image tokens is $\sum_l \rho_l^2 = 3452$. During inference, we use KV-cache for computational saving~\cite{qu2025visualautoregressivemodelingimage, var, hart_iclr25}. 

\subsection{Experimental Setup}
\begin{figure*}[t]
\centering
\begin{minipage}{0.3\textwidth}
\centering \includegraphics[width=0.3\linewidth]{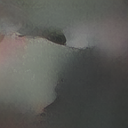}
\end{minipage}
\begin{minipage}{0.3\textwidth}
\centering \includegraphics[width=0.3\linewidth]{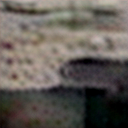}
\end{minipage}
\begin{minipage}{0.3\textwidth}
\centering \includegraphics[width=0.3\linewidth]{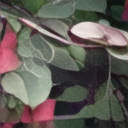}
\end{minipage}


\begin{minipage}{0.3\textwidth}
\centering \includegraphics[width=0.6\linewidth]{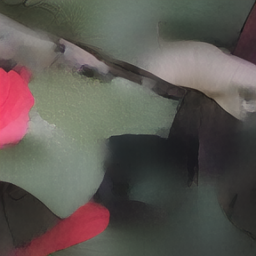}
\end{minipage}
\begin{minipage}{0.3\textwidth}
\centering \includegraphics[width=0.6\linewidth]{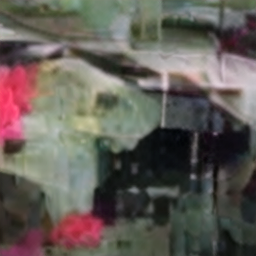}
\end{minipage}
\begin{minipage}{0.3\textwidth}
\centering \includegraphics[width=0.6\linewidth]{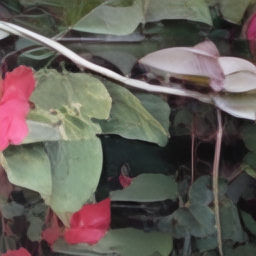}
\end{minipage}

\begin{minipage}{0.3\textwidth}
\centering \includegraphics[width=\linewidth]{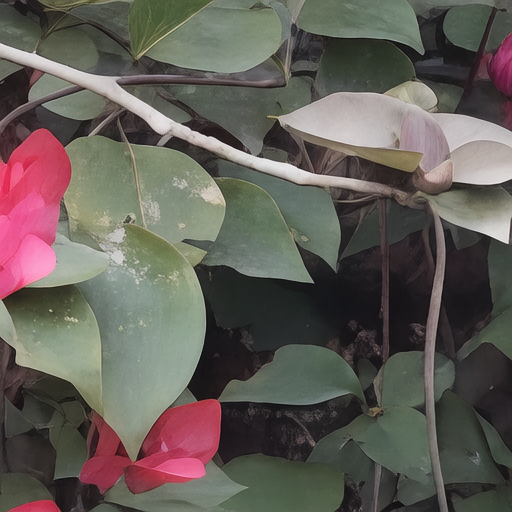}
\end{minipage}
\begin{minipage}{0.3\textwidth}
\centering \includegraphics[width=\linewidth]{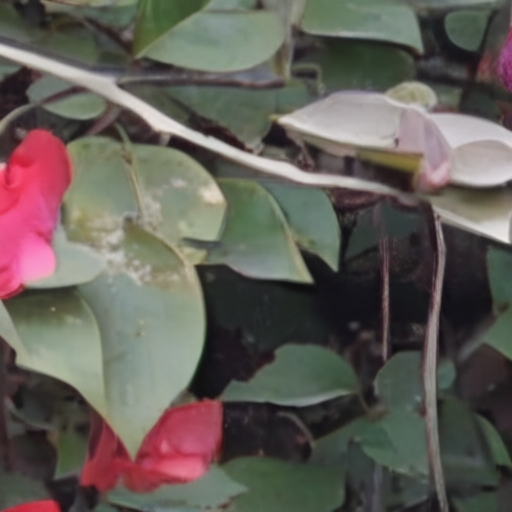}
\end{minipage}
\begin{minipage}{0.3\textwidth}
\centering \includegraphics[width=\linewidth]{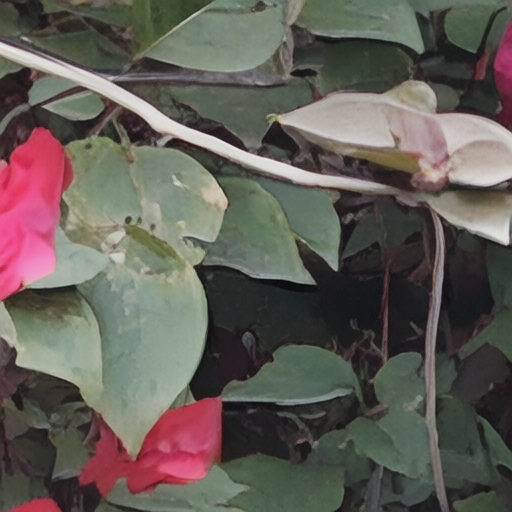}
\end{minipage}
\begin{minipage}{0.3\textwidth}

\end{minipage}

\begin{minipage}{0.3\textwidth}
\centering
(a)
\end{minipage}
\begin{minipage}{0.3\textwidth}
\centering
(b)
\end{minipage}
\begin{minipage}{0.3\textwidth}
\centering
(c)
\end{minipage}
\caption{Multi-scale SR evaluation of (a) VARSR, (b) Baseline, and (c) Our proposed approach. The baseline is trained using our RQVAE, but by generating the ground-truth sequences without hierarchical tokenization. Top-to-bottom images correspond to output of the models at scales $128\times128$, $256\times256$, $512\times512$, respectively, corresponding to scale factors of $\times1$, $\times2$ and $\times4$. For VARSR we represent outputs at $144$ and $288$, respectively, which are the closest to the target outputs using their sequence of scales $(1, 2, 3, 4, 6, 9, 13, 18, 24, 32)$. Better viewed with zoom. \vspace{-4pt}
}
\label{fig:qualitative_1}
\end{figure*}

\paragraph{Training datasets.} We adopt standard training datasets used for the $\times4$ ISR task and use a combination of DIV2K \cite{div}, DIV8K\cite{div8k}, Flickr2k \cite{flickr2k}, OST \cite{ost} and a subset of 10K images from FFHQ training set \cite{ffhq}. To generate the LR-HR pairs for training, we use the Real-ESRGAN \cite{realesrgan_iccvw21} degradation pipeline. 

\paragraph{Testing datasets.} For evaluation, we follow recent ISR work, e.g., \cite{stablesr,hu2025improvingautoregressivevisualgeneration} and test on one synthetic dataset and two real ones. Specifically, we use the synthetic dataset made of 3K LR-HR ($128\rightarrow512$) pairs synthesized from the DIV2K validation set using the Real-ESRGAN degradation pipeline. For the real datasets, we use $128\times128$ center crops from the RealSR \cite{cai2019toward}, DRealSR \cite{wei2020component} datasets.

\paragraph{Baselines.} We compare to a sample from GAN-based methods, diffusion-based methods, and also the closely related recent VARSR work. Importantly, given that VARSR relies on a big AR model (VAR-d24) and was trained on a much larger dataset that is not publicly available, we also retrain it using the same backbone model adopted in our work (VAR-d16) and using the same training datasets, for fair comparison. 

\paragraph{Evaluation metrics.} We evaluate using standard reference-based metrics, including PSNR, SSIM, LPIPS, and FID where applicable.

\subsection{Results}

\paragraph{RQVAE reconstruction.} We first evaluate the RQVAE reconstruction capabilities at different scales. We compare our finetuned hierarchical RQVAE to the initial checkpoint without (i.e., Switti) and with (i.e., Switti*) our tokenization algorithm described in \cref{alg:enc}. We show in \cref{tab:rqvae_metrics} that our finetuned RQVAE surpasses alternatives in all metrics. While a small and expected degradation occurs at higher scales due to the quantization, we can see that our tokenization algorithm allows for decoding all scales with high fidelity. 
\label{ssec:results}

\begin{figure*}[t]
\centering 
\begin{minipage}{0.25\textwidth}
\includegraphics[width=\linewidth]{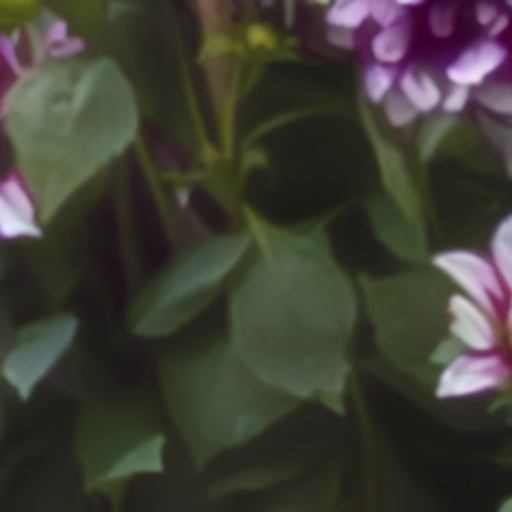}
\end{minipage}
\begin{minipage}{0.25\textwidth}
\includegraphics[width=\linewidth]{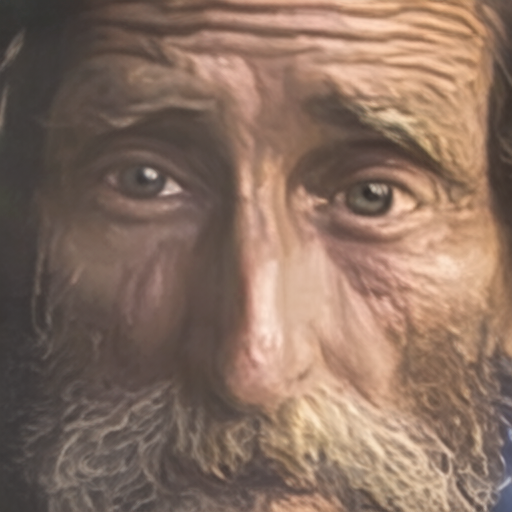}
\end{minipage}
\begin{minipage}{0.25\textwidth}
\includegraphics[width=\linewidth]{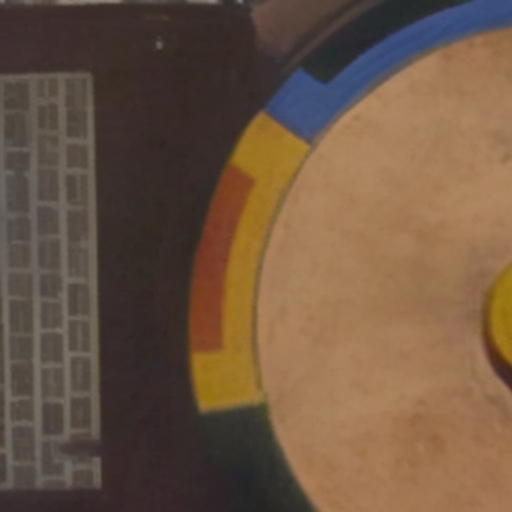}
\end{minipage}



\begin{minipage}{0.25\textwidth}
\includegraphics[width=\linewidth]{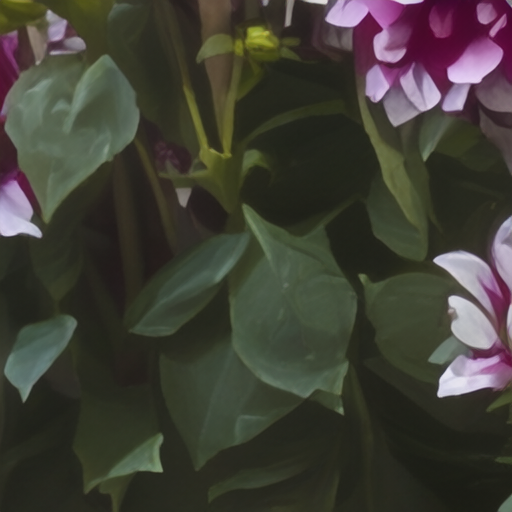}
\end{minipage}
\begin{minipage}{0.25\textwidth}
\includegraphics[width=\linewidth]{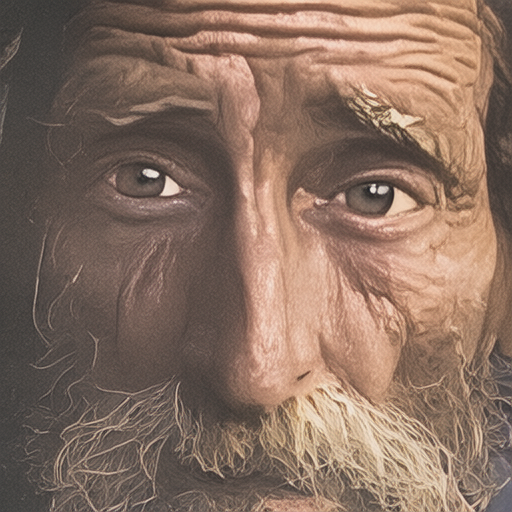}
\end{minipage}
\begin{minipage}{0.25\textwidth}
\includegraphics[width=\linewidth]{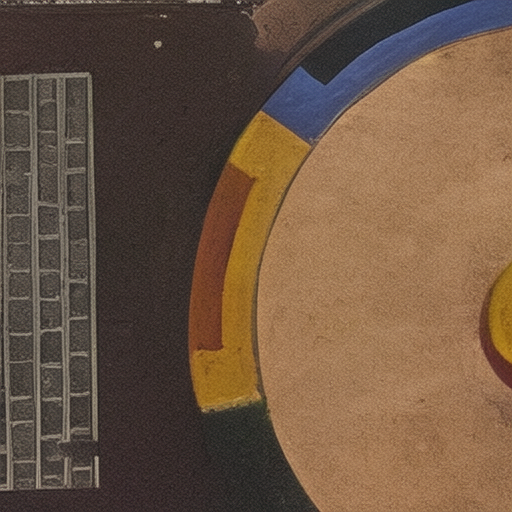}
\end{minipage}



\caption{Qualitative results: (Top) \textbf{without} and (Bottom) \textbf{with} our proposed DPO-based regularization. We can see the role of the regularization term in sharpening results.} 
\label{fig:qualitative_2}
\end{figure*}

\begin{table*}[t!]
\scriptsize
    \begin{subtable}[b]{0.40\linewidth}
        \centering
            \begin{tabular*}{\linewidth}{c | c | c c c }
			\toprule
		\textbf{Method}	& \textbf{Metric} & \textbf{128}  & \textbf{256} & \textbf{512}   \\ 
			\midrule
            \multirow{2}{*}{Switti} & PSNR$\uparrow$ & - & - & 24.17 \\
            & LPIPs$\downarrow$  & - & - & 0.115\\ \midrule
            \multirow{2}{*}{Switti*} & PSNR$\uparrow$  & 17.86 & 21.43 & 23.52 \\
            & LPIPs$\downarrow$ & 0.285 & 0.159 & 0.144\\ \midrule
            \multirow{2}{*}{Ours} & PSNR$\uparrow$ & 20.10 & 22.34 & 24.81 \\
            & LPIPs$\downarrow$ & 0.184 & 0.143 & 0.124 \\
 
               \bottomrule

            \end{tabular*}
            \caption{} \vspace{-0.cm}\label{tab:rqvae_metrics}
    \end{subtable} \hfill 
    \begin{subtable}[b]{0.55\linewidth}
        \centering
            \begin{tabular*}{\linewidth}{c | c | c | c c c }
			\cmidrule{1-6}
		\textbf{Dataset} & \textbf{Method}	& \textbf{Metric} & \textbf{128}  & \textbf{256} & \textbf{512}   \\ 
			\cmidrule{1-6}
            
    \multirow{4}{*}{RealSR} &        \multirow{2}{*}{w/o DPO} & PSNR$\uparrow$  & 20.56 & 23.09 & 25.72 \\
          &  & LPIPs$\downarrow$ & 0.347 & 0.309 & 0.310\\   \cmidrule{2-6}
          &  \multirow{2}{*}{with DPO} & PSNR$\uparrow$ & 22.09 & 24.41 & 26.39 \\
          &  & LPIPs$\downarrow$ & 0.1996 & 0.2357 & 0.2762 \\ \cmidrule{1-6}

          \multirow{4}{*}{DRealSR} &        \multirow{2}{*}{w/o DPO} & PSNR$\uparrow$  & 23.03 & 26.38 & 28.61 \\
          &  & LPIPs$\downarrow$ & 0.311 & 0.311 & 0.335\\   \cmidrule{2-6}
          &  \multirow{2}{*}{with DPO} & PSNR$\uparrow$ & 25.26 & 27.65 & 29.46 \\
          &  & LPIPs$\downarrow$ & 0.1871 & 0.2435 & 0.2957 \\ 
 
               \cmidrule{1-6} 

        \end{tabular*}
        \caption{} \label{tab:dpo_vs_nodpo}
    \end{subtable}
    \hfill
    \caption{a) Reconstruction metrics of different RQVAEs on ImageNet-512~\cite{imagenet} validation set. Switti$*$ denotes results using Switti's checkpoint and the tokenization of \cref{alg:enc} \emph{before} finetuning. b) Ablation study showcasing the contribution of our proposed DPO regularization.
    \vspace{-20pt} }
\end{table*}

\paragraph{ISR at different scales.} Next we evaluate the capacity of our proposed model to upsample images at different scales. In Table~\ref{tab:resos} we compare the results for VARSR against our proposed pipeline for $128$, $256$ and $512$ resolution. To further showcase the importance of our proposed tokenization scheme for the VAR, we train a \textit{baseline} using the very same RQVAE we used for our experiments, but \textit{without the proposed tokenization during VAR training and no DPO regularization}. We additionally show in \cref{fig:qualitative_1} how our method is the only one capable of reconstructing the intermediate scales properly. 

\paragraph{Role of DPO regularization.} We also demonstrate the contribution of our proposed regularization term. We show both numerically in \cref{tab:dpo_vs_nodpo}, and qualitatively in \cref{fig:qualitative_2}, that such regularization helps the model predict sequences that lead to sharper results.

\begin{table*}[t]
  \centering
  \resizebox{0.95\textwidth}{!}{
  \centering
  \begin{tabular}{c|c|ccc|ccc|ccc}
    \toprule
    \multirow{2}{*}{Dataset} & \multirow{2}{*}{Metrics} & \multicolumn{3}{c|}{128} & \multicolumn{3}{c|}{256}& \multicolumn{3}{c}{512} \\
    &  & VARSR & Baseline & \textbf{Ours} & VARSR & Baseline & \textbf{Ours} & VARSR & Baseline & \textbf{Ours} \\
    \midrule
    \multirow{4}{*}{\textit{DIV2K-Val}}
    & PSNR$\uparrow$     & 18.58 & 17.61 & \textbf{21.95} & 20.19 &19.55 & \textbf{24.04} &23.91 & \textbf{24.69} & 24.20 \\
    & SSIM$\uparrow$     & 0.4086 & 0.3811 & \textbf{0.5949} & 0.5010  &0.4868 & \textbf{0.6656} &0.5980 & \textbf{0.6470}& 0.6226\\
    & LPIPS$\downarrow$  &  0.5501 & 0.6530  & \textbf{0.2092} & 0.4086 & 0.5667& \textbf{0.2347} & \textbf{0.3260}& 0.5021& 0.3488\\
    \midrule
    \multirow{4}{*}{\textit{RealSR}}
    & PSNR$\uparrow$     & 18.97 &  17.08& \textbf{22.09} &20.07 & 18.62& \textbf{24.41}& 24.61& 26.11 & \textbf{26.39}\\
    & SSIM$\uparrow$     & 0.4251 & 0.3824 & \textbf{0.6283} & 0.5258&0.4777 & \textbf{0.7252}& 0.7169& 0.7564 & \textbf{0.7593}\\
    & LPIPS$\downarrow$  &0.6178  & 0.6863 & \textbf{0.1996} &0.4504 & 0.4912&\textbf{0.2357} &0.3504 & 0.3110 & \textbf{0.2762}\\
   
    \midrule
    \multirow{4}{*}{\textit{DRealSR}}
    & PSNR$\uparrow$     & 21.56 & 19.97 & \textbf{25.26} &22.96 &22.10 & \textbf{27.65}& 28.16& 29.04 & \textbf{29.46}\\
    & SSIM$\uparrow$     & 0.5431 & 0.4990 & \textbf{0.7043} &0.6399 & 0.6025& \textbf{0.7858} &0.7652 & \textbf{0.8236} & 0.8185\\
    & LPIPS$\downarrow$  & 0.5650 & 0.6362 & \textbf{0.1871} & 0.4345& 0.4563& \textbf{0.2435}& 0.3541& 0.3221 & \textbf{0.2957}\\
    
  \bottomrule
  \end{tabular}
  }
  \caption{Comparison at different scales. Best results are highlighted in bold.   }\label{tab:resos}
  
  \vspace{-20pt} 
\end{table*}

\begin{figure*}[t]
\centering
\begin{minipage}{0.15\textwidth}
\includegraphics[width=\linewidth]{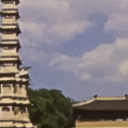}
\end{minipage}
\begin{minipage}{0.15\textwidth}
\includegraphics[width=\linewidth]{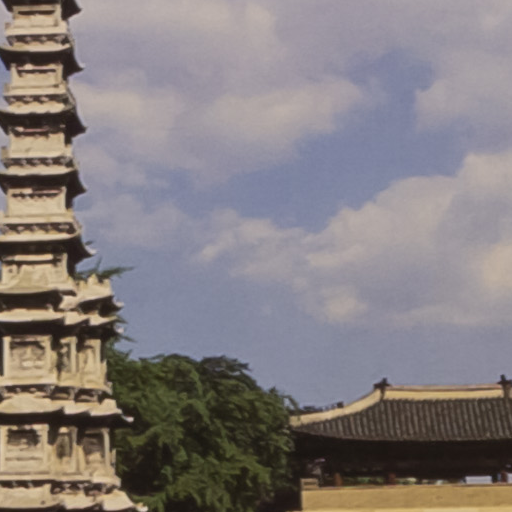}
\end{minipage}
\begin{minipage}{0.15\textwidth}
\includegraphics[width=\linewidth]{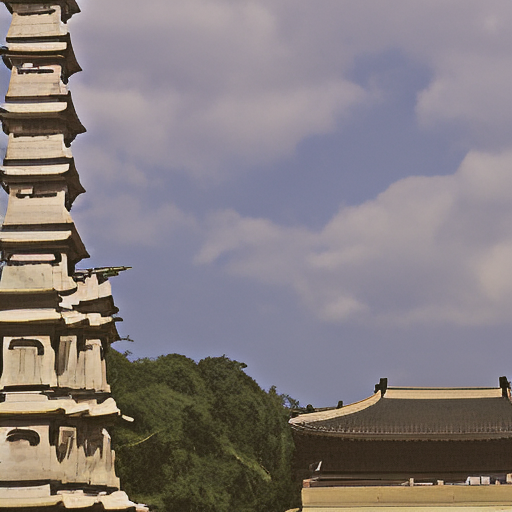}
\end{minipage}
\begin{minipage}{0.15\textwidth}
\includegraphics[width=\linewidth]{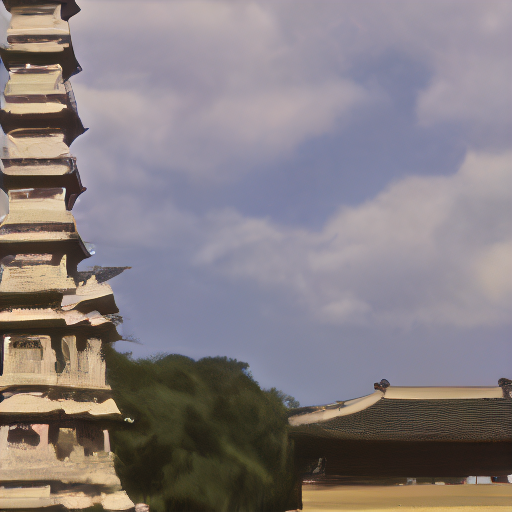}
\end{minipage}
\begin{minipage}{0.15\textwidth}
\includegraphics[width=\linewidth]{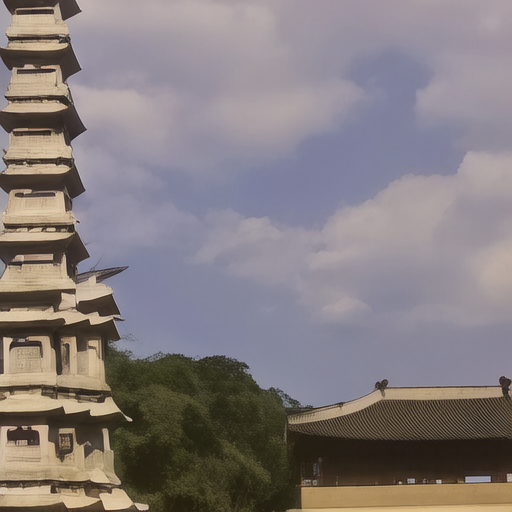}
\end{minipage}
\begin{minipage}{0.15\textwidth}
\includegraphics[width=\linewidth]{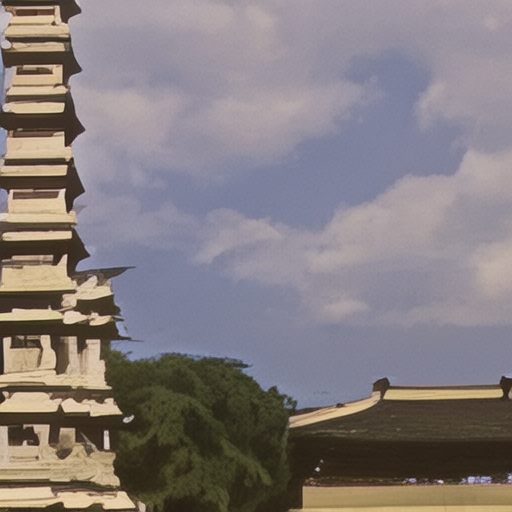}
\end{minipage}



\begin{minipage}{0.15\textwidth}
\includegraphics[width=\linewidth]{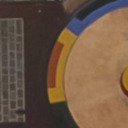}
\end{minipage}
\begin{minipage}{0.15\textwidth}
\includegraphics[width=\linewidth]{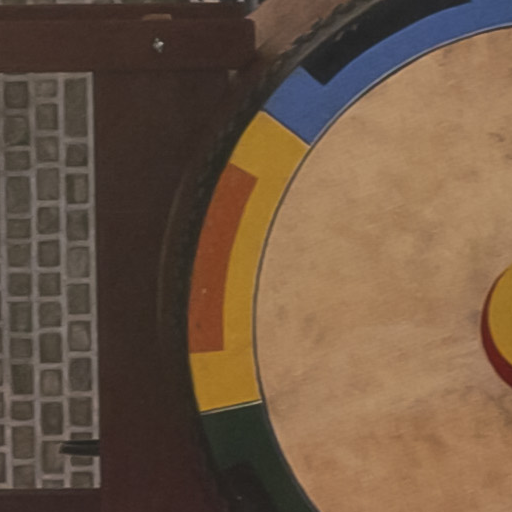}
\end{minipage}
\begin{minipage}{0.15\textwidth}
\includegraphics[width=\linewidth]{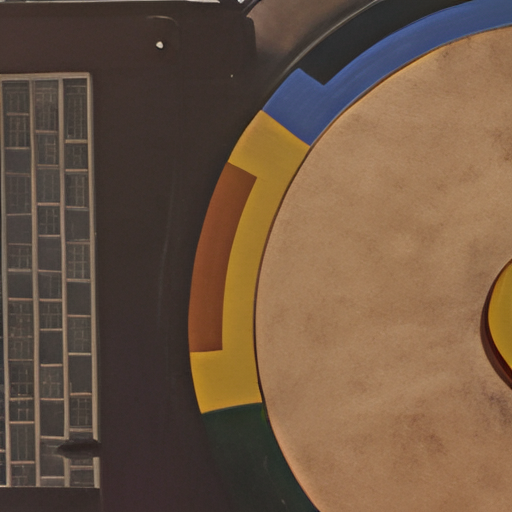}
\end{minipage}
\begin{minipage}{0.15\textwidth}
\includegraphics[width=\linewidth]{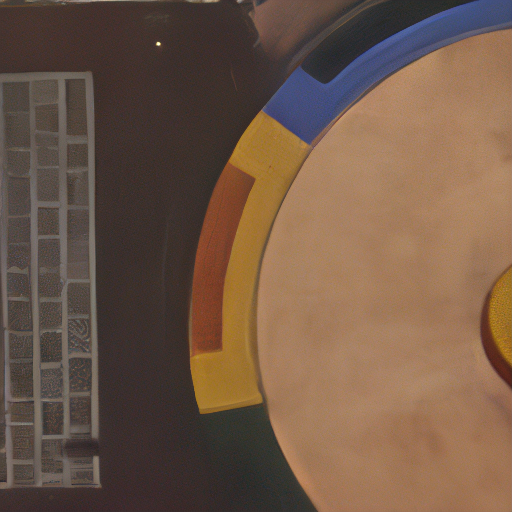}
\end{minipage}
\begin{minipage}{0.15\textwidth}
\includegraphics[width=\linewidth]{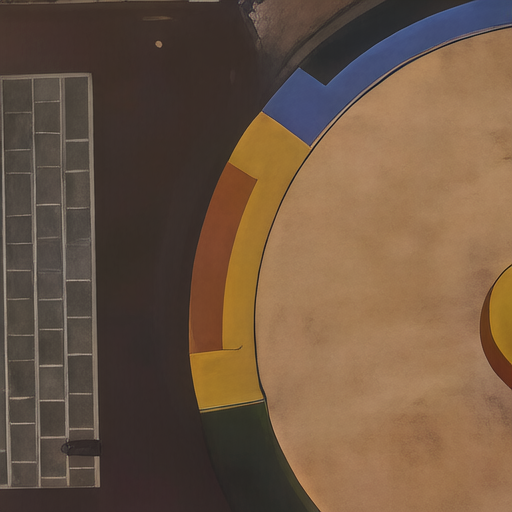}
\end{minipage}
\begin{minipage}{0.15\textwidth}
\includegraphics[width=\linewidth]{isma/isma/multiscale/ours/512/Canon_022_LR4.png}
\end{minipage}

\begin{minipage}{0.15\textwidth}
\includegraphics[width=\linewidth]{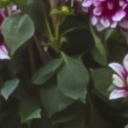}
\end{minipage}
\begin{minipage}{0.15\textwidth}
\includegraphics[width=\linewidth]{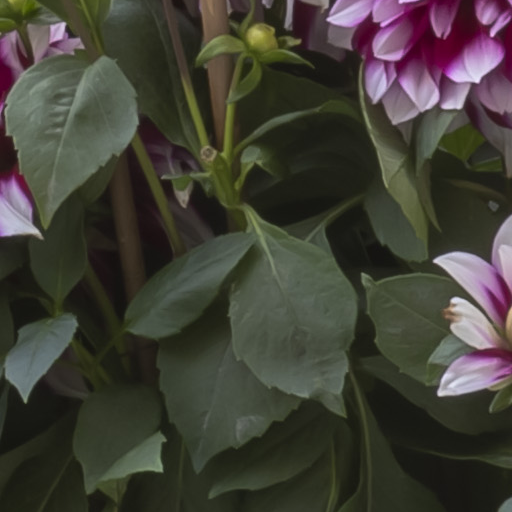}
\end{minipage}
\begin{minipage}{0.15\textwidth}
\includegraphics[width=\linewidth]{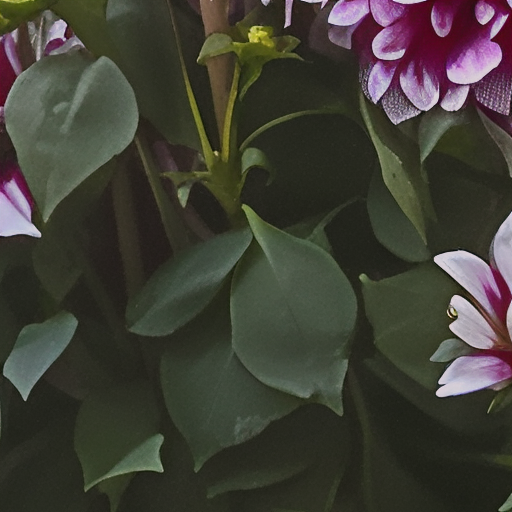}
\end{minipage}
\begin{minipage}{0.15\textwidth}
\includegraphics[width=\linewidth]{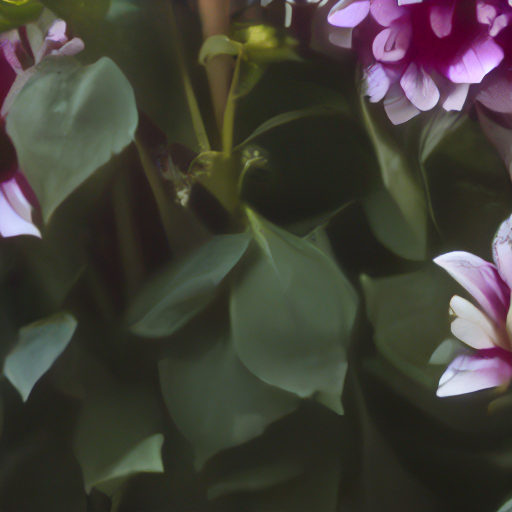}
\end{minipage}
\begin{minipage}{0.15\textwidth}
\includegraphics[width=\linewidth]{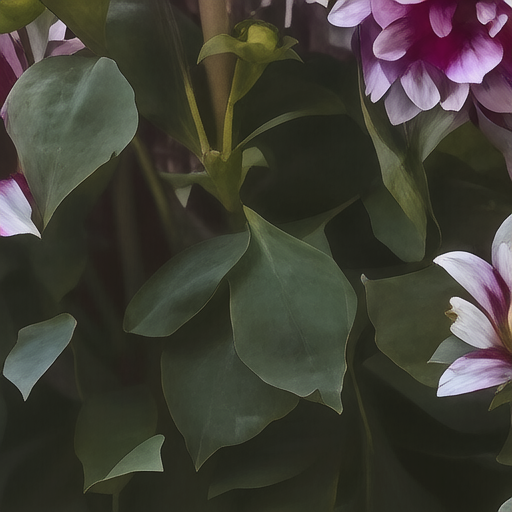}
\end{minipage}
\begin{minipage}{0.15\textwidth}
\includegraphics[width=\linewidth]{isma/isma/multiscale/ours/512/Nikon_043_LR4.png}
\end{minipage}

\begin{minipage}{0.15\textwidth}
\includegraphics[width=\linewidth]{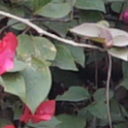}
\end{minipage}
\begin{minipage}{0.15\textwidth}
\includegraphics[width=\linewidth]{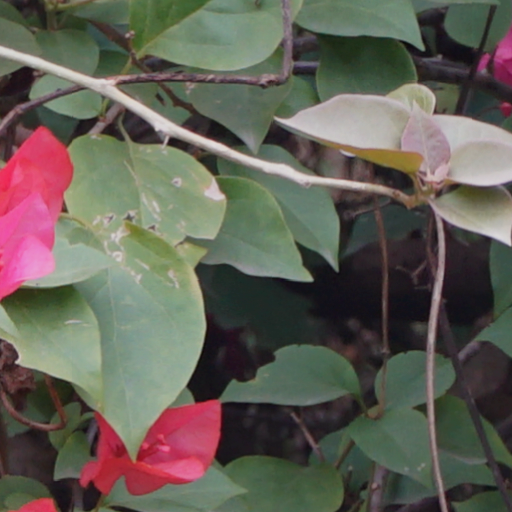}
\end{minipage}
\begin{minipage}{0.15\textwidth}
\includegraphics[width=\linewidth]{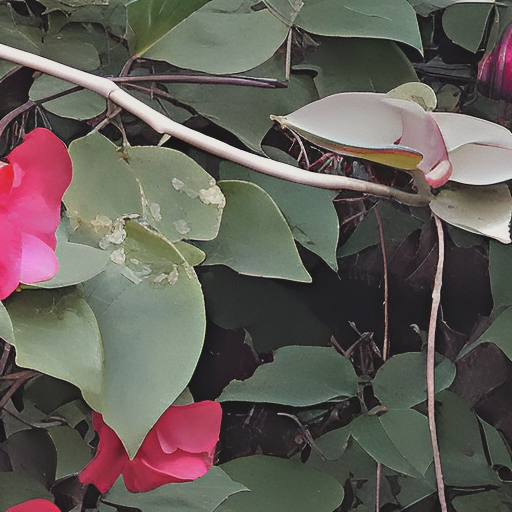}
\end{minipage}
\begin{minipage}{0.15\textwidth}
\includegraphics[width=\linewidth]{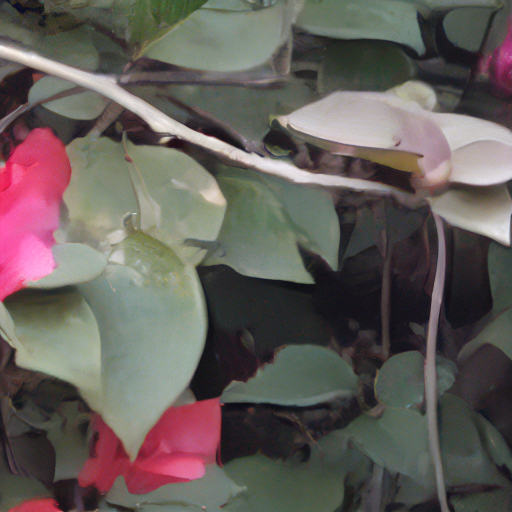}
\end{minipage}
\begin{minipage}{0.15\textwidth}
\includegraphics[width=\linewidth]{isma/isma/multiscale/varsr/512/sony_120_x1.png}
\end{minipage}
\begin{minipage}{0.15\textwidth}
\includegraphics[width=\linewidth]{isma/isma/multiscale/ours/512/sony_120_x1.png}
\end{minipage}

\begin{minipage}{0.15\textwidth}
\includegraphics[width=\linewidth]{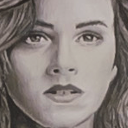}
\end{minipage}
\begin{minipage}{0.15\textwidth}
\includegraphics[width=\linewidth]{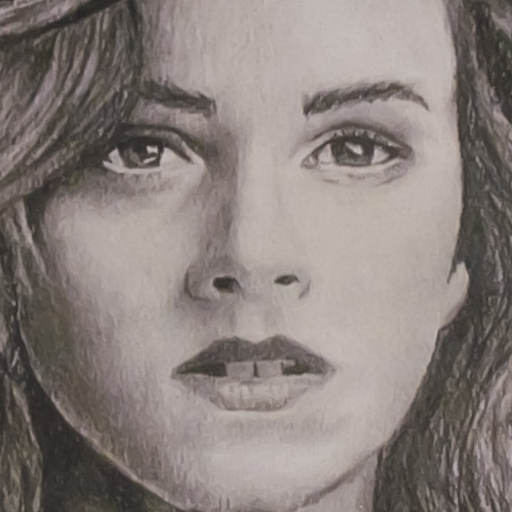}
\end{minipage}
\begin{minipage}{0.15\textwidth}
\includegraphics[width=\linewidth]{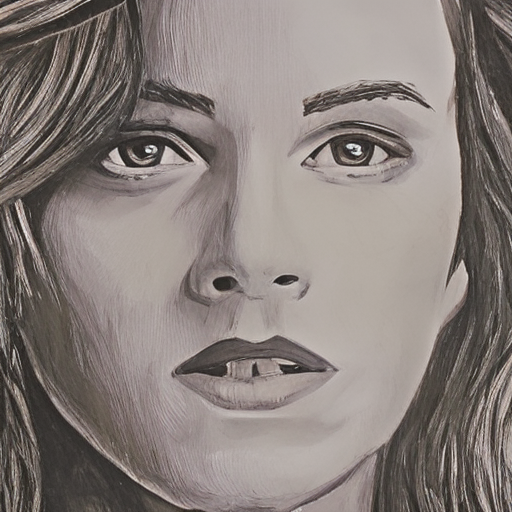}
\end{minipage}
\begin{minipage}{0.15\textwidth}
\includegraphics[width=\linewidth]{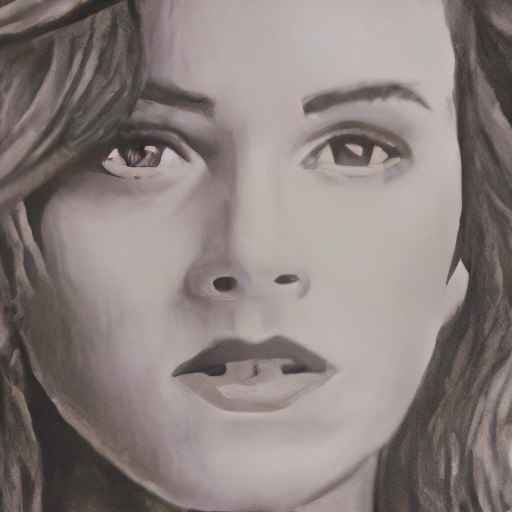}
\end{minipage}
\begin{minipage}{0.15\textwidth}
\includegraphics[width=\linewidth]{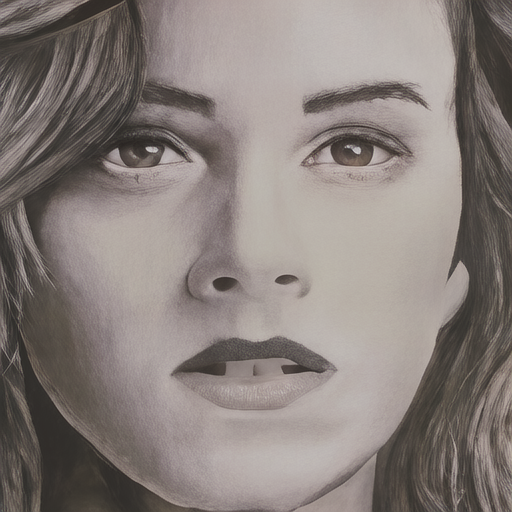}
\end{minipage}
\begin{minipage}{0.15\textwidth}
\includegraphics[width=\linewidth]{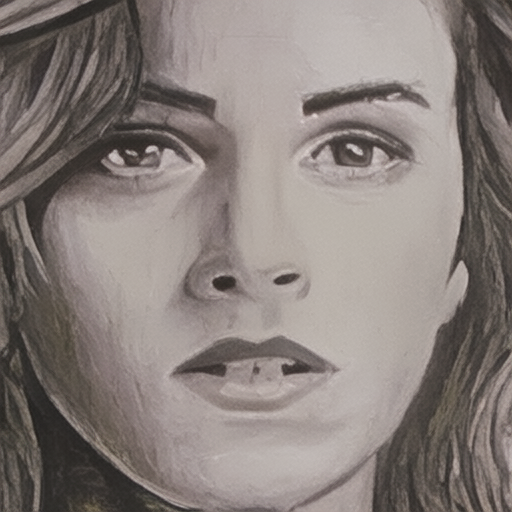}
\end{minipage}
\begin{minipage}{0.15\textwidth}

\end{minipage}

\begin{minipage}{0.15\textwidth}
\centering
(a)
\end{minipage}
\begin{minipage}{0.15\textwidth}
\centering
(b)
\end{minipage}
\begin{minipage}{0.15\textwidth}
\centering
(c)
\end{minipage}
\begin{minipage}{0.15\textwidth}
\centering
(d)
\end{minipage}
\begin{minipage}{0.15\textwidth}
\centering
(e)
\end{minipage}
\begin{minipage}{0.15\textwidth}
\centering
(f)
\end{minipage}
\caption{Qualitative results. (a) Input LR (upsampled to target resolution); (b) Ground truth; (c) StableSR; (d) Resshift; (e) VARSR; (f) Ours. Zoom in for better view.
}
\label{fig:qualitative_sota}
\vspace{-15pt}
\end{figure*}

\paragraph{Comparison with SOTA} We compare our proposed approach against the original VARSR and the retrained variant using the same model size and training datasets as ours. We also, compare to other GAN-based  (BSRGAN~\cite{zhang2021designing}, Real-ESRGAN~\cite{realesrgan_iccvw21}, SwinIR~\cite{liang2021swinir}) and Diffusion-based (LDM~\cite{rombach2022high}, StableSR~\cite{stablesr}, ResShift~\cite{yue2023resshift}) ISR methods. The \textbf{quantitative results} shown in Table~\ref{tab:headings} and \textbf{qualitative comparison} shown in \cref{fig:qualitative_sota} clearly demonstrate the effectiveness of our approach. Notably, ResShift is the closest diffusion-based model in size to ours with $\sim 200M$ params, and VARSR-d16 is the most fair comparison to ours in terms of approach, model size, and training data. Comparison to these methods speaks decisively in favor of our method. In contrast, StableSR and VARSR are similarly sized models with about 1B params, with VARSR benefiting from training on a much larger dataset. Despite being significantly smaller than larger baselines and being trained on small standard SR datasets, our method remains competitive against all baselines.

\begin{table*}[t]
  \centering
  \resizebox{0.99\textwidth}{!}{
  
  \centering
  \begin{tabular}{c|c|ccc|ccccc|ccc}
    \toprule
    \multirow{2}{*}{Dataset} & \multirow{2}{*}{Metrics} & \multicolumn{3}{c|}{GAN-based} & \multicolumn{5}{c|}{Diffusion-based}& \multicolumn{3}{c}{AR-based} \\
    & & BSRGAN & Real-ESR & SwinIR & LDM & StableSR & DiffBIR & SeeSR & ResShift& VARSR & VARSR-d16 & \textbf{Ours} \\
    \midrule
    \multirow{4}{*}{\textit{DIV2K-Val}}
    & PSNR$\uparrow$     & \textcolor{red}{24.42}& \textcolor{blue}{24.30}& 23.77& 21.66& 23.26& 23.49& 23.56& 21.75& 23.91 & 23.14&24.17 \\
    & SSIM$\uparrow$     & 0.6164& \textcolor{red}{0.6324}&0.6186& 0.4752& 0.5670& 0.5568& 0.5981&0.5422&0.5980 & 0.5791& \textcolor{blue}{0.625}\\
    & LPIPS$\downarrow$  & 0.3511&0.3267& 0.3910& 0.4887& \textcolor{red}{0.3228}& 0.3638& 0.3283& 0.4284&\textcolor{blue}{0.3260} &0.4947& 0.358\\
    
    & FID$\downarrow$    & 50.99& 44.34& 44.45& 55.04& \textcolor{red}{28.32}& 34.55&  \textcolor{blue}{28.89}&55.77&35.51 &45.96& 32.48\\

    \midrule
    \multirow{3}{*}{\textit{RealSR}}
    & PSNR$\uparrow$     & \textcolor{red}{26.38}& 25.68& 25.88& 25.66& 24.69& 24.94& 25.31&26.31& 24.61 & 23.79& \textcolor{blue}{25.91}\\
    & SSIM$\uparrow$     & \textcolor{blue}{0.7651}& 0.7614& \textcolor{red}{0.7671}& 0.6934& 0.7090& 0.6664& 0.7284&0.7421& 0.7169 &0.6472& 0.756 \\
    & LPIPS$\downarrow$  & \textcolor{blue}{0.2656}& 0.2710& \textcolor{red}{0.2614}& 0.3367& 0.3003& 0.3485& 0.2993&0.3460& 0.3504 &0.4130& 0.267\\

    \midrule
    \multirow{3}{*}{\textit{DRealSR}}
    & PSNR$\uparrow$     & \textcolor{blue}{28.70} & 28.61& 28.20& 27.78& 27.87& 26.57& 28.13& 28.46&28.16&27.30 & \textcolor{red}{28.76}\\
    & SSIM$\uparrow$     & 0.8028& \textcolor{blue}{0.8052}& 0.7983& 0.7152&  0.7427& 0.6516& 0.7711& 0.7673&0.7652 &0.7487& \textcolor{red}{0.814} \\
    & LPIPS$\downarrow$  & 0.2858& \textcolor{blue}{0.2819}& 0.2830& 0.3745& 0.3333& 0.4537& 0.3142&0.4006&0.3541 &0.4088& \textcolor{red}{0.280}\\
    
  \bottomrule
  \end{tabular}
  }
  \caption{Comparison with SOTA methods on synthetic and real-world benchmarks.  \textcolor{red}{Red} and \textcolor{blue}{blue} colors represent best and second-best results, respectively. 
  }\label{tab:headings}
  
  \vspace{-5mm}
\end{table*}

\section{Related Work}

\paragraph{Image Super Resolution} Existing methods for super resolution~\cite{survey} often tackle the problem of blind recovery, i.e. with unknown degradation parameters. Because blind ISR requires paired data~\cite{dong_eccv14,wang_iccv15,kim_cvpr16}, some works aim at developing a degradation pipeline that is the closest to what real data offers. The most common degradation pipeline nowadays is RealESRGAN~\cite{realesrgan_iccvw21}. In terms of optimization and architecture design, we can split ISR methods into GAN-based~\cite{realesrgan_iccvw21, gansr, Gu_2020_CVPR} and Diffusion-based~\cite{stablesr, yonos_eccv24, codi_arxiv23, edge-sd-sr}. However, while diffusion models provide visually pleasing outputs, their fidelity to the LR image can be poor, e.g., hallucinating texture or details. 

\paragraph{Autoregressive Image Generation} Visual Auto-regressive (VAR) models have very recently challenged the dominance of diffusion models for image generation, e.g.~\cite{var,li_neurips24,switti,fan2025fluid}. Among these, next-scale auto-regressive image models~\cite{var,hart_iclr25,switti,li_neurips24,fan2025fluid} are particularly suited for ISR. They typically rely on a variant of image tokenization~\cite{van2017neural,esser2021taming} based on residual quantization~\cite{lee2022autoregressive}, and are trained to iteratively regress the residuals across a sequence of monotonically increasing resolutions while conditioning on all previous scales. The training scheme offers ideal alignment between pre-training and downstream (ISR) settings. However, to date, only VARSR~\cite{qu2025visualautoregressivemodelingimage} has leveraged next-scale AR models for the ISR problem. Furthermore, it did so by directly applying the VAR formulation, also inheriting its limitations~\cite{var}. In this paper, we advance VARSR by enabling multi-scale image reconstruction, with a small model that is on par with the 1B VARSR model.

\section{Conclusion}
In this paper, we advance the newly proposed paradigm of applying VAR to the task of ISR. By introducing a hierarchical quantization approach, we enable VAR to decode multiple resolutions while performing the next-scale prediction, leading to semantically aligned intermediate results, and demonstrating the preference for such tokenization even for the target of fixed-scale super resolution. Importantly, we tackle multi-scale ISR using a simple yet effective training strategy with a DPO-based regularization term. Our model attains state-of-the-art results with a small, 310M parameter transformer. 

{
    \small
    \bibliographystyle{plain}

    \bibliography{main}

\begin{thebibliography}{10}

\bibitem{div}
Eirikur Agustsson and Radu Timofte.
\newblock {NTIRE} 2017 challenge on single image super-resolution: Dataset and study.
\newblock In {\em IEEE Conference on Computer Vision and Pattern Recognition - Workshops}, 2017.

\bibitem{cai2019toward}
Jianrui Cai, Hui Zeng, Hongwei Yong, Zisheng Cao, and Lei Zhang.
\newblock Toward real-world single image super-resolution: A new benchmark and a new model.
\newblock In {\em IEEE International Conference on Computer Vision}, 2019.

\bibitem{imagenet}
Jia Deng, Wei Dong, Richard Socher, Li-Jia Li, Kai Li, and Li~Fei-Fei.
\newblock Imagenet: A large-scale hierarchical image database.
\newblock In {\em IEEE Conference on Computer Vision and Pattern Recognition}, 2009.

\bibitem{dong_eccv14}
Chao Dong, Chen~Change Loy, Kaiming He, and Xiaoou Tang.
\newblock Learning a deep convolutional network for image super-resolution.
\newblock In {\em European Conference on Computer Vision}, 2014.

\bibitem{esser2021taming}
Patrick Esser, Robin Rombach, and Bjorn Ommer.
\newblock Taming transformers for high-resolution image synthesis.
\newblock In {\em IEEE Conference on Computer Vision and Pattern Recognition}, 2021.

\bibitem{fan2025fluid}
Lijie Fan, Tianhong Li, Siyang Qin, Yuanzhen Li, Chen Sun, Michael Rubinstein, Deqing Sun, Kaiming He, and Yonglong Tian.
\newblock Fluid: Scaling autoregressive text-to-image generative models with continuous tokens.
\newblock In {\em International Conference on Learning Representations}, 2025.

\bibitem{vector_quantization}
R.~Gray.
\newblock Vector quantization.
\newblock {\em IEEE ASSP Magazine}, 1984.

\bibitem{Gu_2020_CVPR}
Jinjin Gu, Yujun Shen, and Bolei Zhou.
\newblock Image processing using multi-code {GAN} prior.
\newblock In {\em IEEE Conference on Computer Vision and Pattern Recognition}, 2020.

\bibitem{div8k}
Shuhang Gu, Andreas Lugmayr, Martin Danelljan, Manuel Fritsche, Julien Lamour, and Radu Timofte.
\newblock {DIV8K: DIVerse 8K} resolution image dataset.
\newblock In {\em IEEE International Conference on Computer Vision - Workshops}, 2019.

\bibitem{Infinity}
Jian Han, Jinlai Liu, Yi~Jiang, Bin Yan, Yuqi Zhang, Zehuan Yuan, Bingyue Peng, and Xiaobing Liu.
\newblock Infinity: Scaling bitwise autoregressive modeling for high-resolution image synthesis.
\newblock In {\em IEEE Conference on Computer Vision and Pattern Recognition}, 2025.

\bibitem{hu2025improvingautoregressivevisualgeneration}
Teng Hu, Jiangning Zhang, Ran Yi, Jieyu Weng, Yabiao Wang, Xianfang Zeng, Zhucun Xue, and Lizhuang Ma.
\newblock Improving autoregressive visual generation with cluster-oriented token prediction, 2025.

\bibitem{ffhq}
Tero Karras, Samuli Laine, and Timo Aila.
\newblock A style-based generator architecture for generative adversarial networks.
\newblock In {\em IEEE Conference on Computer Vision and Pattern Recognition}, 2019.

\bibitem{kim_cvpr16}
Jiwon Kim, Jung~Kwon Lee, and Kyoung~Mu Lee.
\newblock Accurate image super-resolution using very deep convolutional networks.
\newblock In {\em IEEE Conference on Computer Vision and Pattern Recognition}, 2016.

\bibitem{kuznetsova2020open}
Alina Kuznetsova, Hassan Rom, Neil Alldrin, Jasper Uijlings, Ivan Krasin, Jordi Pont-Tuset, Shahab Kamali, Stefan Popov, Matteo Malloci, Alexander Kolesnikov, et~al.
\newblock The open images dataset v4: Unified image classification, object detection, and visual relationship detection at scale.
\newblock {\em International Journal on Computer Vision}, 2020.

\bibitem{gansr}
Christian Ledig, Lucas Theis, Ferenc Huszar, Jose Caballero, Andrew Cunningham, Alejandro Acosta, Andrew~P. Aitken, Alykhan Tejani, Johannes Totz, Zehan Wang, and Wenzhe Shi.
\newblock Photo-realistic single image super-resolution using a generative adversarial network.
\newblock In {\em IEEE Conference on Computer Vision and Pattern Recognition}, 2017.

\bibitem{lee2022autoregressive}
Doyup Lee, Chiheon Kim, Saehoon Kim, Minsu Cho, and Wook-Shin Han.
\newblock Autoregressive image generation using residual quantization.
\newblock In {\em IEEE Conference on Computer Vision and Pattern Recognition}, 2022.

\bibitem{li_neurips24}
Tianhong Li, Yonglong Tian, He~Li, Mingyang Deng, and Kaiming He.
\newblock Autoregressive image generation without vector quantization.
\newblock In {\em Neural Information Processing Systems}, 2024.

\bibitem{liang2021swinir}
Jingyun Liang, Jiezhang Cao, Guolei Sun, Kai Zhang, Luc Van~Gool, and Radu Timofte.
\newblock Swinir: Image restoration using swin transformer.
\newblock {\em arXiv preprint arXiv:2108.10257}, 2021.

\bibitem{survey}
Anran Liu, Yihao Liu, Jinjin Gu, Yu~Qiao, and Chao Dong.
\newblock Blind image superresolution: A survey and beyond.
\newblock In {\em arXiv preprint arXiv:2107.03055}, 2021.

\bibitem{adamw}
Ilya Loshchilov and Frank Hutter.
\newblock Decoupled weight decay regularization.
\newblock In {\em International Conference on Learning Representations}, 2019.

\bibitem{ma2024star}
Xiaoxiao Ma, Mohan Zhou, Tao Liang, Yalong Bai, Tiejun Zhao, Huaian Chen, and Yi~Jin.
\newblock {STAR}: Scale-wise text-to-image generation via auto-regressive representations.
\newblock {\em arXiv preprint arXiv:2406.10797}, 2024.

\bibitem{codi_arxiv23}
Kangfu Mei, Mauricio Delbracio, Hossein Talebi, Zhengzhong Tu, Vishal~M Patel, and Peyman Milanfar.
\newblock {CoDi}: Conditional diffusion distillation for higher-fidelity and faster image generation.
\newblock {\em IEEE Conference on Computer Vision and Pattern Recognition}, 2024.

\bibitem{edge-sd-sr}
Mehdi Noroozi, Isma Hadji, Victor Escorcia, Anestis Zaganidis, Brais Martinez, and Georgios Tzimiropoulos.
\newblock Edge-sd-sr: Low latency and parameter efficient on-device super-resolution with stable diffusion via bidirectional conditioning.
\newblock In {\em IEEE Conference on Computer Vision and Pattern Recognition}, 2025.

\bibitem{yonos_eccv24}
Mehdi Noroozi, Isma Hadji, Brais Martinez, Adrian Bulat, and Georgios Tzimiropoulos.
\newblock You only need one step: Fast super-resolution with stable diffusion via scale distillation.
\newblock In {\em European Conference on Computer Vision}, 2024.

\bibitem{qu2025visualautoregressivemodelingimage}
Yunpeng Qu, Kun Yuan, Jinhua Hao, Kai Zhao, Qizhi Xie, Ming Sun, and Chao Zhou.
\newblock Visual autoregressive modeling for image super-resolution.
\newblock In {\em International Conference on Machine Learning}, 2025.

\bibitem{radford2019language}
Alec Radford, Jeffrey Wu, Rewon Child, David Luan, Dario Amodei, Ilya Sutskever, et~al.
\newblock Language models are unsupervised multitask learners.
\newblock {\em OpenAI blog}, 1(8):9, 2019.

\bibitem{rafailov2023direct}
Rafael Rafailov, Archit Sharma, Eric Mitchell, Christopher~D Manning, Stefano Ermon, and Chelsea Finn.
\newblock Direct preference optimization: Your language model is secretly a reward model.
\newblock In {\em Neural Information Processing Systems}, 2023.

\bibitem{rombach2022high}
Robin Rombach, Andreas Blattmann, Dominik Lorenz, Patrick Esser, and Bj{\"o}rn Ommer.
\newblock High-resolution image synthesis with latent diffusion models.
\newblock In {\em Proceedings of the IEEE/CVF conference on computer vision and pattern recognition}, pages 10684--10695, 2022.

\bibitem{su_roformer}
Jianlin Su, Murtadha Ahmed, Yu~Lu, Shengfeng Pan, Wen Bo, and Yunfeng Liu.
\newblock Roformer: Enhanced transformer with rotary position embedding.
\newblock {\em Neurocomputing}, 2024.

\bibitem{sun2024autoregressive}
Peize Sun, Yi~Jiang, Shoufa Chen, Shilong Zhang, Bingyue Peng, Ping Luo, and Zehuan Yuan.
\newblock Autoregressive model beats diffusion: Llama for scalable image generation.
\newblock {\em arXiv preprint arXiv:2406.06525}, 2024.

\bibitem{hart_iclr25}
Haotian Tang, Yecheng Wu, Shang Yang, Enze Xie, Junsong Chen, Junyu Chen, Zhuoyang Zhang, Han Cai, Yao Lu, and Song Han.
\newblock {HART:} efficient visual generation with hybrid autoregressive transformer.
\newblock {\em International Conference on Learning Representations}, 2025.

\bibitem{var}
Keyu Tian, Yi~Jiang, Zehuan Yuan, Bingyue Peng, and Liwei Wang.
\newblock Visual autoregressive modeling: Scalable image generation via next-scale prediction.
\newblock In {\em Neural Information Processing Systems}, 2024.

\bibitem{flickr2k}
Radu Timofte, Eirikur Agustsson, Luc~Van Gool, MingHsuan Yang, and Lei Zhang.
\newblock {NTIRE} 2017 challenge on single image super-resolution: Methods and results.
\newblock In {\em IEEE Conference on Computer Vision and Pattern Recognition - Workshops}, 2017.

\bibitem{van2017neural}
Aaron Van Den~Oord, Oriol Vinyals, and Koray Kavukcuoglu.
\newblock Neural discrete representation learning.
\newblock {\em Neural Information Processing Systems}, 2017.

\bibitem{switti}
Anton Voronov, Denis Kuznedelev, Mikhail Khoroshikh, Valentin Khrulkov, and Dmitry Baranchuk.
\newblock Switti: Designing scale-wise transformers for text-to-image synthesis.
\newblock In {\em IEEE Conference on Computer Vision and Pattern Recognition}, 2025.

\bibitem{stablesr}
Jianyi Wang, Zongsheng Yue, Shangchen Zhou, Kelvin~C.K. Chan, and Chen~Change Loy.
\newblock Exploiting diffusion prior for real-world image super-resolution.
\newblock {\em International Journal on Computer Vision}, 2024.

\bibitem{realesrgan_iccvw21}
Xintao Wang, Liangbin Xie, Chao Dong, and Ying Shan.
\newblock {Real-ESRGAN}: Training real-world blind super-resolution with pure synthetic data.
\newblock In {\em IEEE International Conference on Computer Vision - Workshops}, 2021.

\bibitem{ost}
Xintao Wang, Ke~Yu, Chao Dong, and Chen~Change Loy.
\newblock Recovering realistic texture in image super-resolution by deep spatial feature transform.
\newblock In {\em IEEE Conference on Computer Vision and Pattern Recognition}, 2018.

\bibitem{wang_iccv15}
Zhaowen Wang, Ding Liu, Jianchao Yang, Wei Han, and Thomas Huang.
\newblock Deep networks for image super-resolution with sparse prior.
\newblock In {\em IEEE International Conference on Computer Vision}, 2015.

\bibitem{wei2020component}
Pengxu Wei, Ziwei Xie, Hannan Lu, Zongyuan Zhan, Qixiang Ye, Wangmeng Zuo, and Liang Lin.
\newblock Component divide-and-conquer for real-world image super-resolution.
\newblock In {\em European Conference on Computer Vision}, 2020.

\bibitem{yue2023resshift}
Zongsheng Yue, Jianyi Wang, and Chen~Change Loy.
\newblock Resshift: Efficient diffusion model for image super-resolution by residual shifting.
\newblock In {\em Neural Information Processing Systems}, 2023.

\bibitem{zhang2021designing}
Kai Zhang, Jingyun Liang, Luc Van~Gool, and Radu Timofte.
\newblock Designing a practical degradation model for deep blind image super-resolution.
\newblock In {\em IEEE International Conference on Computer Vision}, 2021.

\bibitem{zhang2023adding}
Lvmin Zhang, Anyi Rao, and Maneesh Agrawala.
\newblock Adding conditional control to text-to-image diffusion models.
\newblock In {\em IEEE International Conference on Computer Vision}, 2023.

\bibitem{zhang2018perceptual}
Richard Zhang, Phillip Isola, Alexei~A Efros, Eli Shechtman, and Oliver Wang.
\newblock The unreasonable effectiveness of deep features as a perceptual metric.
\newblock In {\em IEEE Conference on Computer Vision and Pattern Recognition}, 2018.

\end{thebibliography}
}
\section*{Appendix}
\begin{figure*}[ht]
\centering
\begin{minipage}{0.3\textwidth}
\centering \includegraphics[width=0.3\linewidth]{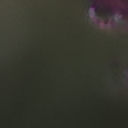}
\end{minipage}
\begin{minipage}{0.3\textwidth}
\centering \includegraphics[width=0.3\linewidth]{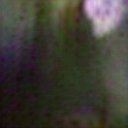}
\end{minipage}
\begin{minipage}{0.3\textwidth}
\centering \includegraphics[width=0.3\linewidth]{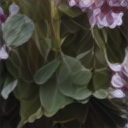}
\end{minipage}


\begin{minipage}{0.3\textwidth}
\centering \includegraphics[width=0.6\linewidth]{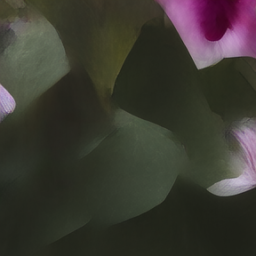}
\end{minipage}
\begin{minipage}{0.3\textwidth}
\centering \includegraphics[width=0.6\linewidth]{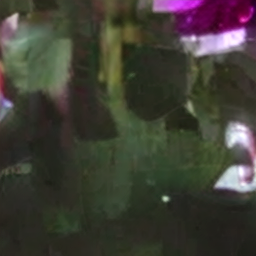}
\end{minipage}
\begin{minipage}{0.3\textwidth}
\centering \includegraphics[width=0.6\linewidth]{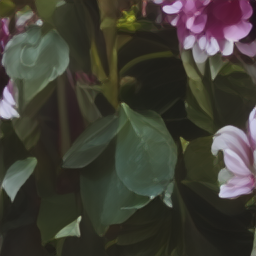}
\end{minipage}

\begin{minipage}{0.3\textwidth}
\centering \includegraphics[width=\linewidth]{isma/isma/multiscale/varsr/512/Nikon_043_LR4.png}
\end{minipage}
\begin{minipage}{0.3\textwidth}
\centering \includegraphics[width=\linewidth]{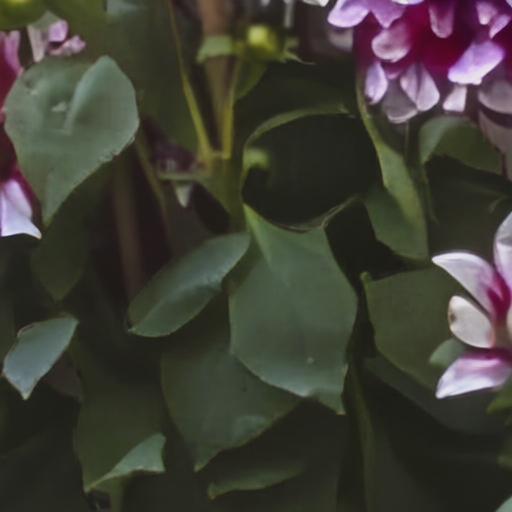}
\end{minipage}
\begin{minipage}{0.3\textwidth}
\centering \includegraphics[width=\linewidth]{isma/isma/multiscale/ours/512/Nikon_043_LR4.png}
\end{minipage}
\begin{minipage}{0.3\textwidth}

\end{minipage}

\begin{minipage}{0.3\textwidth}
\centering
(a)
\end{minipage}
\begin{minipage}{0.3\textwidth}
\centering
(b)
\end{minipage}
\begin{minipage}{0.3\textwidth}
\centering
(c)
\end{minipage}
\caption{Multi-scale SR evaluation of (a) VARSR, (b) Baseline, and (c) Our proposed approach. The baseline is trained using our RQVAE, but by generating the ground-truth sequences without hierarchical tokenization. Top-to-bottom images correspond to output of the models at scales $128\times128$, $256\times256$, $512\times512$, respectively, corresponding to scale factors of $\times1$, $\times2$ and $\times4$. For VARSR we represent outputs at $144$ and $288$, respectively, which are the closest to the target outputs using their sequence of scales $(1, 2, 3, 4, 6, 9, 13, 18, 24, 32)$. Better viewed with zoom. \vspace{-4pt}
}
\label{fig:qualitative_3}
\end{figure*}

\begin{figure*}[t]
\centering
\begin{minipage}{0.95\textwidth}
\includegraphics[width=\linewidth]{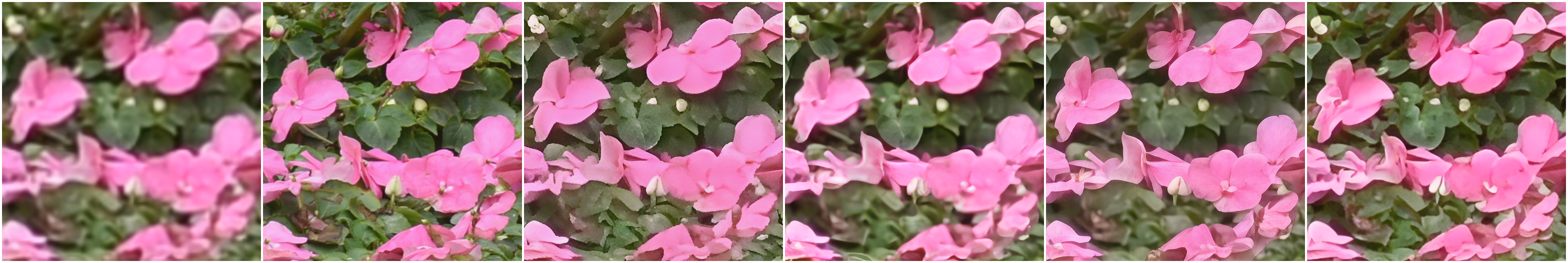}
\end{minipage}
\begin{minipage}{0.95\textwidth}
\includegraphics[width=\linewidth]{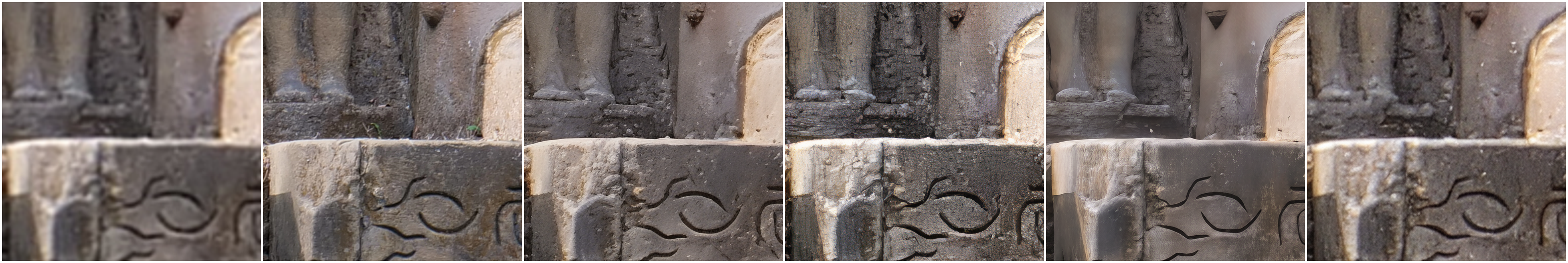}
\end{minipage}
\begin{minipage}{0.95\textwidth}
\includegraphics[width=\linewidth]{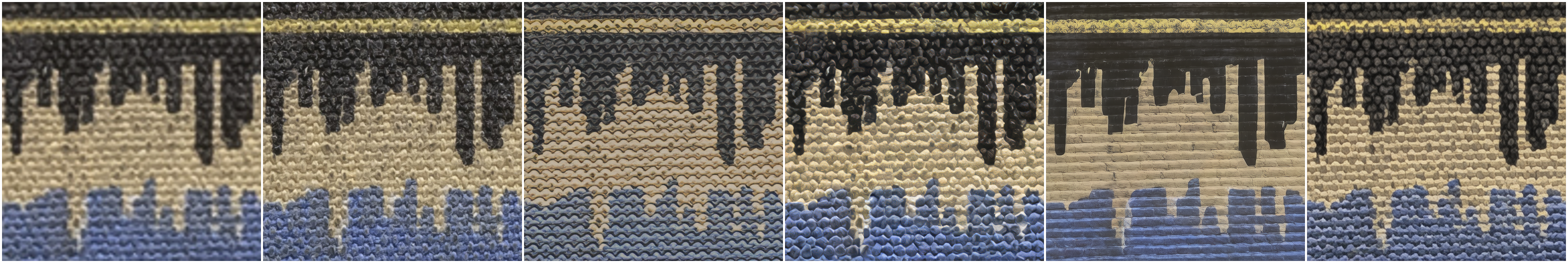}
\end{minipage}
\begin{minipage}{0.95\textwidth}
\includegraphics[width=\linewidth]{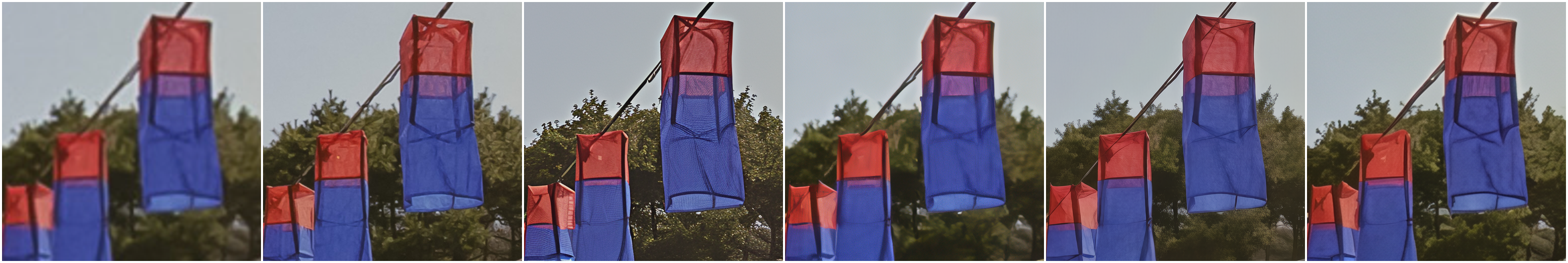}
\end{minipage}

\begin{minipage}{0.15\textwidth}
\centering
(a)
\end{minipage}
\begin{minipage}{0.15\textwidth}
\centering
(b)
\end{minipage}
\begin{minipage}{0.15\textwidth}
\centering
(c)
\end{minipage}
\begin{minipage}{0.15\textwidth}
\centering
(d)
\end{minipage}
\begin{minipage}{0.15\textwidth}
\centering
(e)
\end{minipage}
\begin{minipage}{0.15\textwidth}
\centering
(f)
\end{minipage}
\caption{Qualitative results. (a) Input LR (upsampled to target resolution); (b) Ground truth; (c) StableSR; (d) Resshift; (e) VARSR; (f) Ours. Zoom in for better view.
}
\label{fig:qualitative_sota2}
\vspace{-15pt}
\end{figure*}

In \cref{fig:qualitative_3} we present more examples of the limitations of the baseline and VARSR to produce semantically consistent images across scales. In \cref{fig:qualitative_sota2} we show more qualitative examples of our method compared with state of the art. 

\begin{figure*}[t]
\centering

\begin{minipage}{0.95\textwidth}
\includegraphics[width=\linewidth]{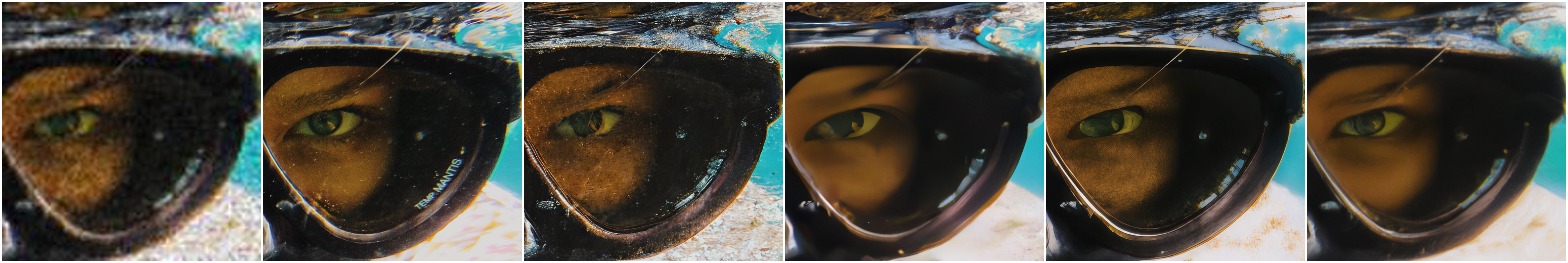}
\end{minipage}
\begin{minipage}{0.95\textwidth}
\includegraphics[width=\linewidth]{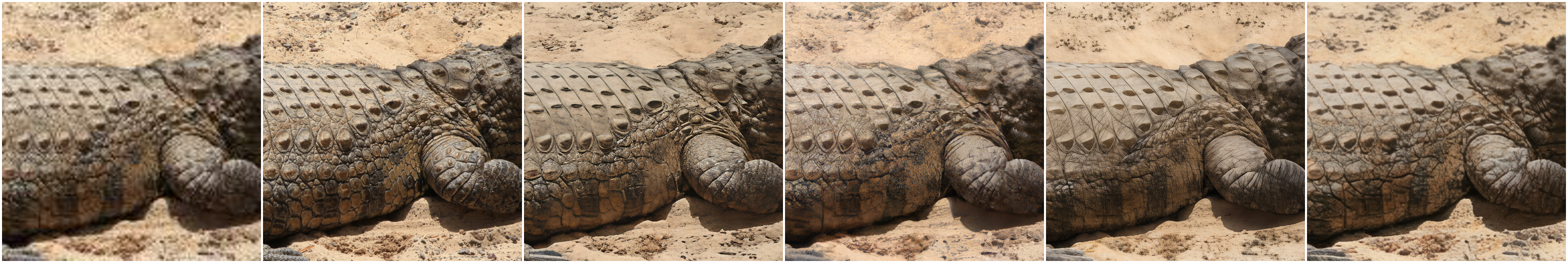}
\end{minipage}

\begin{minipage}{0.95\textwidth}
\includegraphics[width=\linewidth]{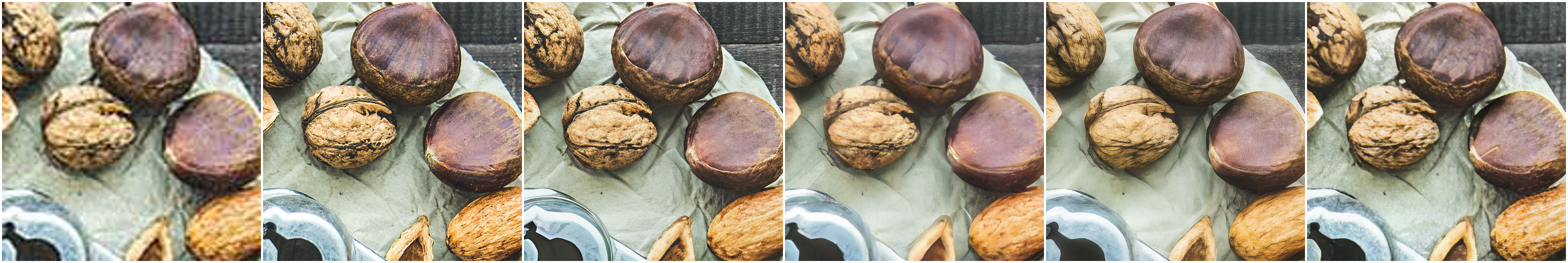}
\end{minipage}
\begin{minipage}{0.95\textwidth}
\includegraphics[width=\linewidth]{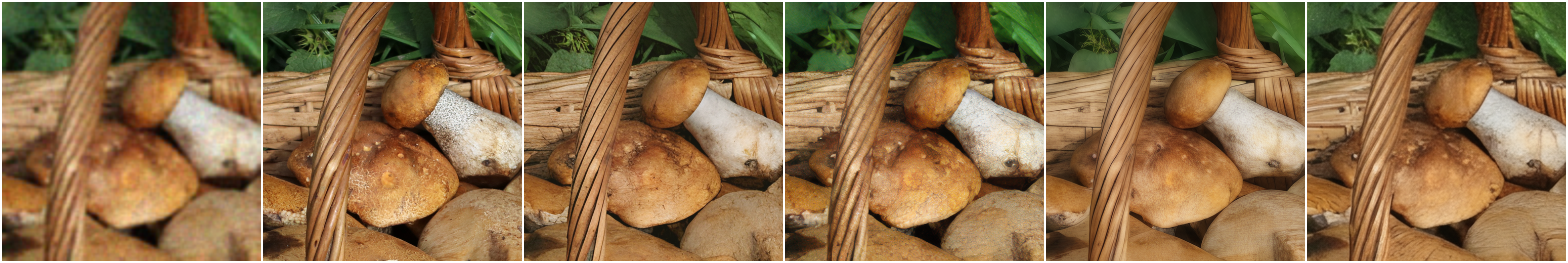}
\end{minipage}

\begin{minipage}{0.15\textwidth}
\centering
(a)
\end{minipage}
\begin{minipage}{0.15\textwidth}
\centering
(b)
\end{minipage}
\begin{minipage}{0.15\textwidth}
\centering
(c)
\end{minipage}
\begin{minipage}{0.15\textwidth}
\centering
(d)
\end{minipage}
\begin{minipage}{0.15\textwidth}
\centering
(e)
\end{minipage}
\begin{minipage}{0.15\textwidth}
\centering
(f)
\end{minipage}
\caption{Qualitative results. (a) Input LR (upsampled to target resolution); (b) Ground truth; (c) StableSR; (d) Resshift; (e) VARSR; (f) Ours. Zoom in for better view.
}
\label{fig:qualitative_sota3}
\vspace{-15pt}
\end{figure*}

\medskip

\newpage





\end{document}